\documentclass[english,american,journal]{IEEEtran}
\usepackage[T1]{fontenc}
\usepackage[utf8]{inputenc}
\usepackage[letterpaper]{geometry}
\usepackage{color}
\usepackage{babel}
\usepackage{array}
\usepackage{multirow}
\usepackage{amsmath}
\usepackage{amssymb}
\usepackage{graphicx}
\usepackage[unicode=true,hyperfootnotes=false]{hyperref}

\makeatletter

\providecommand{\tabularnewline}{\\}

\usepackage[T1]{fontenc}

\usepackage{array}

\usepackage{color}

\usepackage{upquote}

\usepackage[usenames,dvipsnames]{xcolor}
\newcommand{\add}[1]{#1}
\usepackage{listings}
\usepackage[most]{tcolorbox}

\usepackage{caption}
\usepackage{graphics}
\usepackage{placeins}
\usepackage{graphicx, epstopdf}
\usepackage[framed, numbered]{matlab-prettifier}
\usepackage{colortbl}
\usepackage{empheq}
\usepackage{footmisc} %

\definecolor{celadon}{rgb}{0.67, 0.88, 0.69}
\definecolor{hellgelb}{rgb}{1,1,0.85} 
\definecolor{colKeys}{rgb}{0,0,1} 
\definecolor{colIdentifier}{rgb}{0,0,0} 
\definecolor{colComments}{rgb}{0,0.5,0} 
\definecolor{colString}{rgb}{0.81,0.12,0.95}

\definecolor{deepblue}{rgb}{0,0,1}
\definecolor{deepred}{rgb}{0.6,0,0}
\definecolor{deepgreen}{rgb}{0,0.5,0}
\definecolor{blue_light}{RGB}{16,161,239}

\hypersetup{
  colorlinks,
  citecolor=black,
  linkcolor=black,
  urlcolor=blue}

\usepackage[algo2e,ruled,vlined,linesnumbered]{algorithm2e}
\usepackage{subfig}
\usepackage{caption}

\usepackage[dvipsnames]{xcolor}

\usepackage{boldline} 

\usepackage{diagbox}

\usepackage{colortbl}
\usepackage{tikz}
\usepackage{tabularx}
\usepackage{dcolumn, array, booktabs}

\usepackage{siunitx}

\usepackage[noadjust]{cite} %

\@ifundefined{showcaptionsetup}{}{%
 \PassOptionsToPackage{caption=false}{subfig}}
\usepackage{subfig}
\makeatother

\begin{document}
\markboth{In review, October~2020}%
{Shell \MakeLowercase{\textit{et al.}}: Bare Demo of IEEEtran.cls for IEEE Journals}
\title{MADER: Trajectory Planner in Multi-Agent and Dynamic Environments}
\author{Jesus Tordesillas and Jonathan P. How\thanks{The authors are with the Aerospace Controls Laboratory, MIT, 77 Massachusetts
Ave., Cambridge, MA, USA \{\tt{jtorde, jhow}\}@mit.edu}}

\maketitle
\renewcommand{\lstlistingname}{Script}

\newcommand{\tikzcircle}[2][black,fill=red]{\tikz[baseline=0.0ex, line width=0.3mm]\draw[#1] [#1] (0,0.08) circle (0.09);}%

\hyphenation{decentralized generate B-Spline challenging}

\begin{abstract}
This paper presents MADER, a 3D decentralized and asynchronous trajectory
planner for UAVs that generates collision-free trajectories in environments
with static obstacles, dynamic obstacles, and other planning agents. Real-time collision avoidance with other dynamic obstacles or agents is done by performing outer polyhedral representations
of every interval of the trajectories and then including the plane
that separates each pair of polyhedra as a decision variable in the
optimization problem. MADER uses our recently developed MINVO basis (\cite{tordesillas2020minvo}) to obtain outer polyhedral representations with volumes 2.36 and 254.9 times, respectively, smaller than the Bernstein or B-Spline
bases used extensively in the planning literature. 
Our decentralized and asynchronous algorithm guarantees
safety with respect to other agents 
by including their committed trajectories as constraints in the optimization and then executing 
a collision check-recheck scheme. Finally, extensive simulations
in challenging cluttered environments show up to a 33.9\% reduction in the flight time, and a 88.8\% reduction in the number of stops compared to the Bernstein and B-Spline bases, shorter flight distances than centralized approaches, and shorter total times on average than synchronous decentralized approaches. 
\end{abstract}

\IEEEpeerreviewmaketitle

\vspace{0.2cm}
\noindent
\add{
	{\footnotesize\textbf{Acronyms:} UAV (Unmanned Aerial Vehicle), MINVO (Basis that obtains a minimum-volume simplex enclosing a polynomial curve \cite{tordesillas2020minvo}), MADER (Trajectory Planner in Multi-Agent and Dynamic Environments), AABB (Axis-Aligned Bounding Box).}
}

\section*{Supplementary material}%

\noindent
\textbf{Video}: \href{https://youtu.be/aoSoiZDfxGE}{https://youtu.be/aoSoiZDfxGE } \\
\textbf{Code}: \href{https://github.com/mit-acl/mader}{https://github.com/mit-acl/mader}

\section{Introduction and related work}\label{sec:introductionRelWork}

\begin{figure}[t]
\begin{centering}
\subfloat[Circle configuration: 32 agents, 25 static obstacles and 25 dynamic obstacles.\label{fig:Circle32}]{\centering{}\includegraphics[width=1\columnwidth]{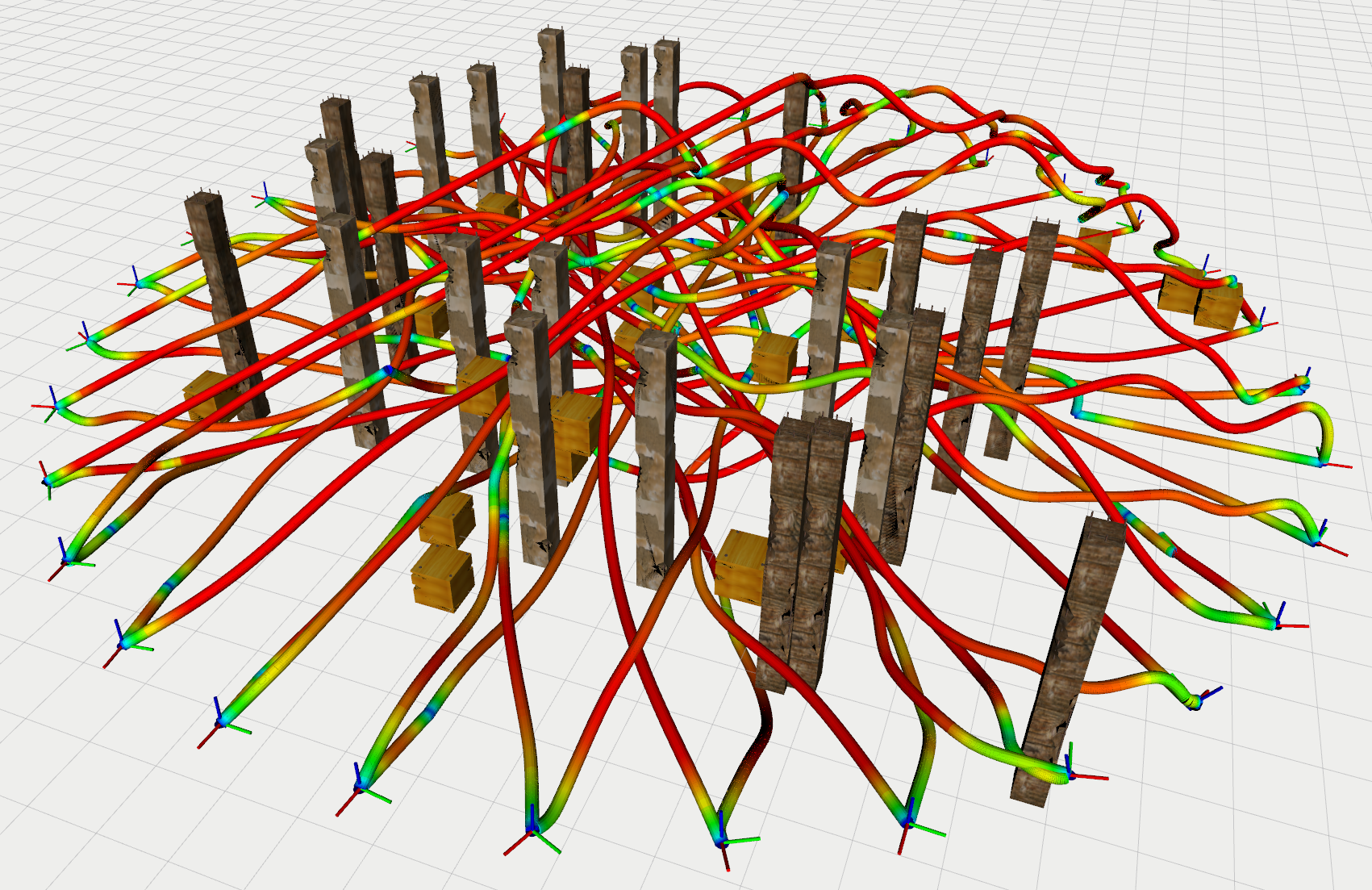}}\hfill{}\subfloat[Sphere configuration: 32 agents, 18 static obstacles and 52 dynamic obstacles. \label{fig:Sphere32}]{\centering{}\includegraphics[width=1\columnwidth]{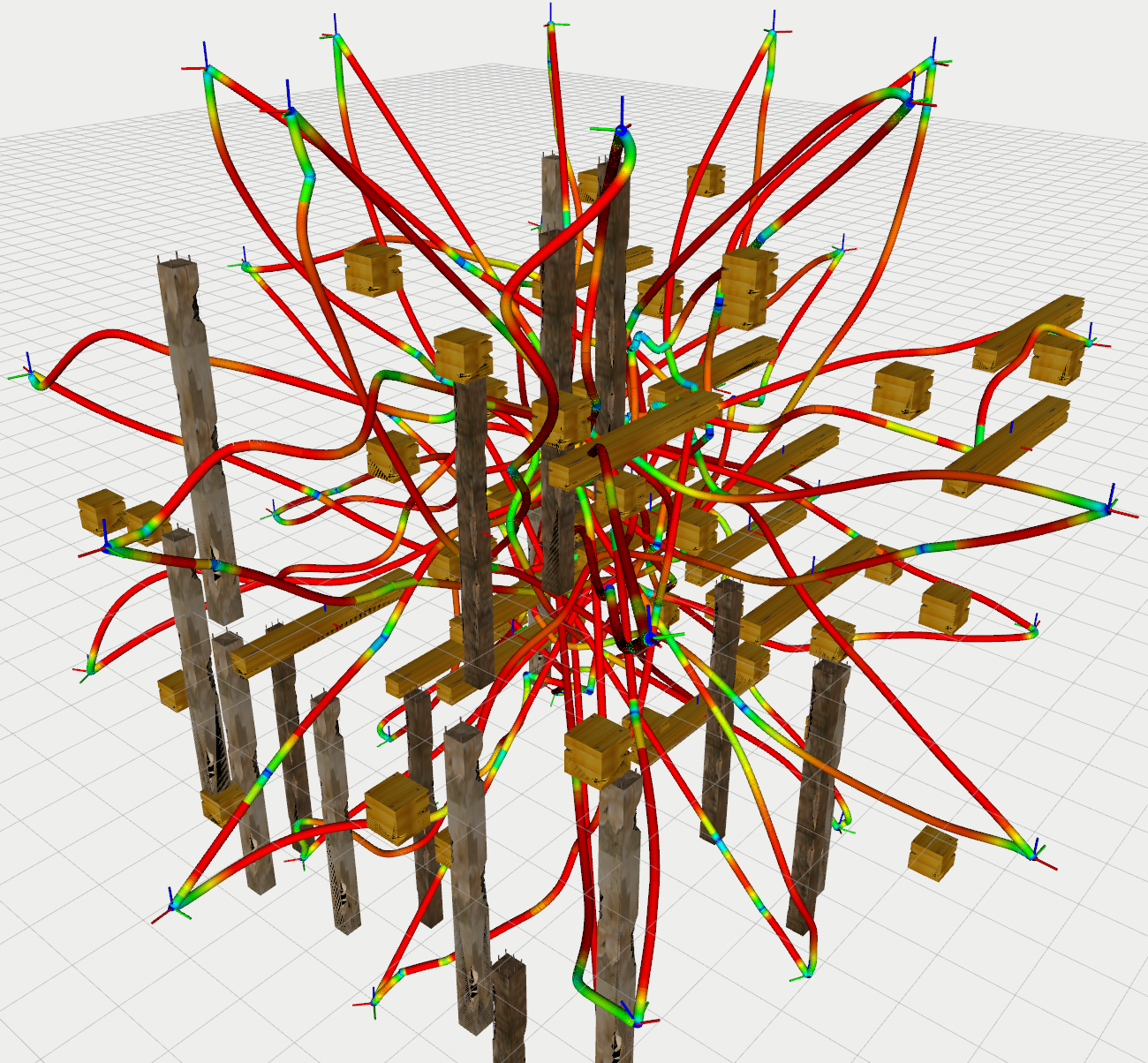}}
\par\end{centering}
\caption{32 agents using MADER to plan trajectories in a decentralized and asynchronous way in an environment with \textbf{dynamic} \textbf{obstacles} (light
brown boxes and horizontal poles), \textbf{static obstacles }(dark
brown pillars) and \textbf{other agents}. \label{fig:32-agents-circle-sphere}}

\vskip-2.3ex
\end{figure}

\IEEEPARstart{W}{hile} efficient and fast UAV trajectory planners
for static worlds have been extensively proposed in the literature
\cite{zhou2020ego,bucki2020rectangular,zhou2020raptor,florence2016integrated,florence2018nanomap,lopez2017aggressive3D,lopez2017aggressivelimitedFOV,tordesillas2018real,tordesillas2019faster,chen2016online,gao2019flying,zhou2019robust,oleynikova2016continuous},
a 3D real-time planner able to handle environments with static obstacles,
dynamic obstacles \emph{and} other planning agents still remains an open
problem (see Fig.~\ref{fig:32-agents-circle-sphere}).

\definecolor{representation_color}{RGB}{179,227,204}
\definecolor{deconfliction_color}{RGB}{245,247,167}
\definecolor{collision_free_color}{RGB}{255,149,149}

\begin{figure}
	\begin{centering}
		\includegraphics[width=0.95\columnwidth]{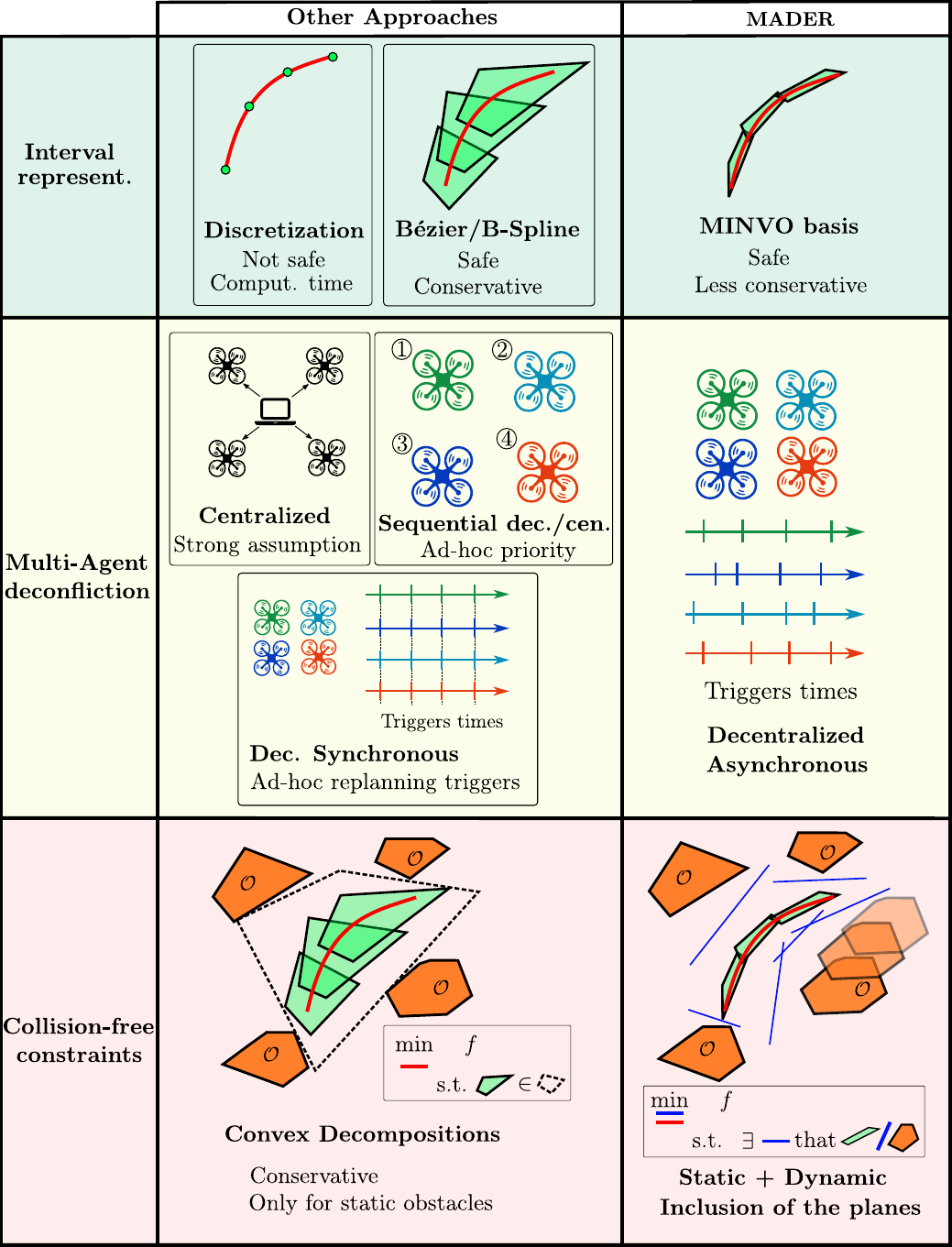}
		\par\end{centering}
	\caption{Contributions of MADER.  \label{fig:contributions} }
	
	\vskip-2ex
\end{figure}

To be able to guarantee safety, the trajectory of the planning agent and the ones of other obstacles/agents need to be encoded in the optimization (see Fig. \ref{fig:contributions}). A common \fcolorbox{black}{representation_color!50}{\textbf{representation}} of this trajectory in the optimization is via points discretized along the trajectory \cite{mellinger2012mixed, potdar2020online, zhu2019chance, szmuk2018real, gao2017quadrotor, lin2020robust}. However, this does not usually guarantee safety between two consecutive discretization points and alleviating that problem by using a fine discretization of the trajectory can lead to a very high computational burden. To reduce this computational burden, polyhedral
outer representations of each interval of the trajectory are extensively
used in the literature, with the added benefit of ensuring safety
at all times (i.e., not just at the discretization points). A common way to obtain this polyhedral outer representation
is via the convex hull of the control points of the Bernstein basis
(basis used by B\'{e}zier curves) or the B-Spline basis \cite{zhou2019robust,tordesillas2019faster,preiss2017trajectory,tang2019real}.
However, these bases do not yield very tight (i.e., with minimum volume) tetrahedra that enclose the curve, leading to conservative results.

MADER addresses this conservatism at its source and leverages our recently developed MINVO basis \cite{tordesillas2020minvo} to obtain control points that generate the $n$-simplex (a tetrahedron for $n=3$) with the minimum volume that completely contains each interval of the curve. Global optimality \add{(in terms of minimum volume)}
of this
tetrahedron obtained by the MINVO basis is guaranteed both in position and velocity space. 

When other agents are present, the \fcolorbox{black}{deconfliction_color!50}{\textbf{deconfliction}} problem between the trajectories also needs to be solved. Most of the state-of-the-art approaches either rely on centralized algorithms \cite{augugliaro2012generation,kushleyev2013towards,wu2019temporal}
and/or on imposing an ad-hoc priority such that an agent only avoids other agents with higher priority %
\cite{robinson2018efficient,park2019efficient,chen2015decoupled, morgan2016swarm, ma2016decentralized}. Some decentralized solutions have also been proposed \cite{luis2019trajectory, chen2015decoupled, liu2018towards}, but they require synchronization between the replans of different agents. The challenge then is how to create a decentralized and asynchronous planner that solves the deconfliction problem and guarantees safety and feasibility for all the agents. 

MADER solves this deconfliction in a decentralized and asynchronous way by including the trajectories other agents have committed to as constraints in the optimization. After the optimization, a collision check-recheck scheme ensures that the trajectory found is still feasible with respect to the trajectories other agents have committed to while the optimization was happening. 

To impose \fcolorbox{black}{collision_free_color!50}{\textbf{collision-free constraints}} in the presence of static obstacles, a common approach is to \emph{first} find convex decompositions of free space and \emph{then} force (in the optimization problem) the outer polyhedral representation of each interval to be inside these convex decompositions \cite{tordesillas2019faster,liu2017planning,deits2015efficient}.
However, this approach can be conservative, especially in cluttered
environments in which the convex decomposition algorithm may not
find a tight representation of the free space. In the presence of dynamic obstacles, these convex decompositions become harder, and likely intractable, due to the extra time dimension. 

To be able to impose collision-free constraints with respect to dynamic obstacles/agents, 
MADER imposes the separation between the polyhedral representations of each trajectory via planes. Moreover, MADER overcomes the conservatism of a convex decomposition (imposed ad-hoc before the optimization) by including a parameterization of these separating planes as decision variables in the optimization problem.
The solver can thus choose the optimal location of these planes to determine collision avoidance. Including this plane parameterization reduces conservatism, but it comes at the expense of creating a nonconvex problem, for which a good initial guess is imperative. For this initial guess, we present a search-based algorithm that handles dynamic environments and obtains both the control points of the trajectory and the planes that separate it from other obstacles/agents.

The contributions of this paper are therefore summarized as follows (see also Fig. \ref{fig:contributions}):
\begin{itemize}
\item Decentralized and asynchronous planning framework that solves the
deconfliction between the agents by imposing as constraints the
trajectories other agents have committed to, and then doing a collision check-recheck scheme to guarantee safety with respect to trajectories other agents have committed to during the optimization time.  
\item Collision-free constraints are imposed by using a novel polynomial basis in trajectory planning: the MINVO basis. In position space, the MINVO basis yields a volume 2.36 and 254.9 times smaller than the extensively-used Bernstein and B-Spline bases, respectively.
\item Formulation of the collision-free constraints with respect to other dynamic obstacles/agents by including the planes that separate the outer polyhedral representations of each interval of every pair of trajectories as decision variables. 
\item Extensive simulations and comparisons with state-of-the-art baselines in cluttered environments. The results show up to a 33.9\% reduction in the flight time, a 88.8\% reduction in the number of stops (compared to Bernstein/B-Spline bases), shorter flight distances than centralized approaches, and shorter total times on average than synchronous decentralized approaches. 
\end{itemize}

\section{Definitions }\label{sec:definitions}

\newcommand{\NextTraj}{\tikz[baseline=0.0ex]\draw [red,thick, dash pattern=on 3pt off 1pt] (0,0.08) -- (0.5,0.08);}

\newcommand{\CurrTraj}{\tikz[baseline=0.0ex]\draw [red,thick] (0,0.08) -- (0.5,0.08);}

\newcommand{\modifiedFig}[1]{\fcolorbox{white}{white}{#1}}
\newcommand{\ff}[1]{f_j^{\text{BS}\rightarrow\text{MV}}(#1)}
\newcommand{\ffQ}{\ff{\mathcal{Q}_{j}^{\text{BS}}}}
\newcommand{\ffDot}{\ff{\cdot}}

\newcommand{\hh}[1]{h_j^{\text{BS}\rightarrow\text{MV}}(#1)}
\newcommand{\hhQ}{\hh{\mathcal{Q}_{j}^{\text{BS}}}}
\newcommand{\hhDot}{\hh{\cdot}}

\begin{table}
\begin{centering}
\caption{Notation used in this paper \label{tab:Notation-used-in}}
\par\end{centering}
\noindent\resizebox{\columnwidth}{!}{%
\begin{centering}
\begin{tabular}{|>{\centering}m{0.14\columnwidth}|>{\raggedright}m{0.86\columnwidth}|} %
\hline 
\textbf{Symbol} & \textbf{\qquad \qquad \qquad \qquad Meaning}\tabularnewline
\hline 
\hline 
$\mathbf{p},\mathbf{v},\mathbf{a},\mathbf{j}$ & Position, Velocity, Acceleration and Jerk, $\in\mathbb{R}^{3}$.\tabularnewline
\hline 
$\mathbf{x}$ & State vector: $\arraycolsep=1.4pt\mathbf{x}:=\left[\begin{array}{ccc}
\mathbf{p}^{T} & \mathbf{v}^{T} & \mathbf{a}^{T}\end{array}\right]^{T}\in\mathbb{R}^{9}$\tabularnewline
\hline 
$m$ & $m+1$ is the number of knots of the B-Spline.\tabularnewline
\hline 
$n$ & $n+1$ is the number of control points of the B-Spline.\tabularnewline
\hline 
$p$ & Degree of the polynomial of each interval of the B-Spline. In this
paper we will use $p=3$. \tabularnewline
\hline 
$J$ & Set that contains the indexes of all the intervals of a B-Spline $J:=\{0,1,...,m-2p-1\}$.\tabularnewline
\hline 
\add{$\xi$} & \add{$\xi:=\text{Number of agents + Number of obstacles}$}\tabularnewline
\hline 
$s$ & Index of the planning agent.\tabularnewline
\hline 
$I$ & Set that contains the indexes of all the obstacles/agents, except the
agent $s$. \add{$I:=\{0,1,...,\xi\}\backslash s$}.\tabularnewline
\hline 
$L$ & $L=\{0,1,...,n\}$.\tabularnewline
\hline 
$l$ & Index of the control point. $l\in L$ for position, $l\in L\backslash\{n\}$
for velocity and $l\in L\backslash\{n-1,n\}$ for acceleration.\tabularnewline
\hline 
$i$ & Index of the obstacle/agent, $i\in I$.\tabularnewline
\hline 
$j$ & Index of the interval, $j\in J$.\tabularnewline
\hline 
\add{$\rho$} & \add{Radius of the sphere that models the agents.}\tabularnewline
\hline 
$B_{i}$\add{, $B_s$}  & \add{$B_{i}$ is the 3D axis-aligned bounding box (AABB) of the shape of the agent/obstacle $i$. For simplicity, we assume that the obstacles do not rotate. Hence, $B_{i}$ does not change for a given obstacle/agent $i$. The AABB of the planning agent is denoted as $B_s$.
}\tabularnewline
\hline 
$\boldsymbol{\eta}_{s}$  & Each entry of $\boldsymbol{\eta}_{s}$ is the length of each side of
the AABB of the planning agent (agent whose index is $s$). \textcolor{black}{I.e., $\arraycolsep=1.0pt\boldsymbol{\eta}_{s}:=2\left[\begin{array}{ccc}
		\rho & \rho & \rho\end{array}\right]^{T}\in\mathbb{R}^{3}$}\tabularnewline
\hline 
$\mathcal{C}_{ij}$ & Set of vertexes of the polyhedron that completely encloses the trajectory
of the obstacle/agent $i$ during the initial and final times of the
interval $j$ of the agent $s$.\tabularnewline
\hline 
$\mathbf{c}$ & Vertex of a polyhedron, $\in\mathbb{R}^{3}$.\tabularnewline
\hline 
$\boldsymbol{q},\boldsymbol{v},\boldsymbol{a}$ & Position, velocity, and acceleration control points, $\in\mathbb{R}^{3}$.\tabularnewline
\hline 
$b$ & Notation for the basis used: MINVO ($b=\text{MV}$), Bernstein ($b=\text{Be}$),
or B-Spline ($b=\text{BS}$).\tabularnewline
\hline 
\add{$\mathcal{Q}_{j}^{b}$}
& \add{Set that contains the 4 position control points of the interval $j$ of the trajectory
of the agent $s$ using the basis $b$. $\mathcal{Q}_{j-1}^{\text{MV}}\cap\mathcal{Q}_{j}^{\text{MV}}=\emptyset$ in general. If $b=\text{Be}$, the last control point of interval $j-1$ is also the first control point of interval $j$. If $b=\text{BS}$, the last 3 control points of interval $j-1$ are also the first 3 control points of interval $j$.
Analogous definition for the set $\mathcal{V}_{j}^{b}$, which contains the three
velocity control points.}
\tabularnewline
\hline 
\add{$\boldsymbol{Q}_{j}^{b}$}
& \add{Matrix whose columns contain the 4 position control points of the interval $j$ of the trajectory of the agent $s$ using the basis $b$. Analogous
definition for the matrix $\arraycolsep=1.4pt\boldsymbol{V}_{j}^{b}$, whose columns are the three velocity control points.}\tabularnewline
\hline 
\add{$\ff{\cdot}$}&\add{Linear function (see Eq. \ref{eq:BS2MV}) such that $\mathcal{Q}_{j}^{\text{MV}}=\ffQ$}\tabularnewline
\hline 
\add{$\hh{\cdot}$}&\add{Linear function (see Eq. \ref{eq:BS2MV}) such that $\mathcal{V}_{j}^{\text{MV}}=\hhQ$}\tabularnewline
\hline 
$\boldsymbol{\pi}_{ij}$ ($\boldsymbol{n}_{ij}$, $\;d_{ij}$) & Plane $\boldsymbol{n}_{ij}^{T}\boldsymbol{x}+d_{ij}=0$ that separates
$\mathcal{C}_{ij}$ from $\mathcal{Q}_{j}^{b}$. \tabularnewline
\hline 
$\boldsymbol{1}$, $\text{abs}\left(\boldsymbol{a}\right)$, $\boldsymbol{a}\le\boldsymbol{b}$,
$\oplus$, $\text{conv}(\cdot)$ & Column vector of ones, element-wise absolute value, element-wise inequality,
Minkowski sum, and convex hull.\tabularnewline
\hline 
\scalebox{1.0}[1.6]{\includegraphics[width=0.15\columnwidth]{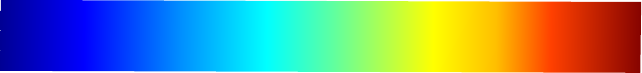}} & Unless otherwise noted, this colormap in the trajectories will represent
the norm of the velocity (blue $0$ m/s and red $v_{\text{max}}$).\tabularnewline
\hline 
\multicolumn{2}{|l|}{Snapshot at $t=t_{1}$ (current time): }\tabularnewline
\multicolumn{2}{|c|}{\includegraphics[width=0.6\columnwidth]{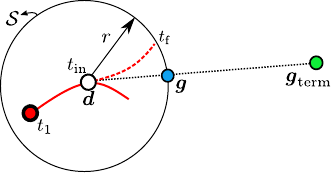}}\tabularnewline
\multicolumn{2}{|l|}{$\boldsymbol{g}_{\text{term}}$ (\tikzcircle[black,fill=green]{15pt})
is the terminal goal, and \tikzcircle[black,fill=red]{15pt} is the
current position of the UAV.}\tabularnewline
\multicolumn{2}{|l|}{\CurrTraj \ is the trajectory the UAV is currently executing. }\tabularnewline
\multicolumn{2}{|l|}{\NextTraj \ is the \add{trajectory} the UAV is currently optimizing, starts
at}\tabularnewline
\multicolumn{2}{|l|}{$t=t_{\text{in}}$ and finishes at $t=t_{\text{f}}$.}\tabularnewline
\multicolumn{2}{|l|}{$\boldsymbol{d}$ (\tikzcircle[black,fill=white]{15pt}) is a point
in \CurrTraj, used as the initial position of \NextTraj.}\tabularnewline
\multicolumn{2}{|l|}{$\mathcal{S}$ is a sphere of radius $r$ around $\boldsymbol{d}$.
\NextTraj \ will be contained in $\mathcal{S}$.}\tabularnewline
\multicolumn{2}{|l|}{$\boldsymbol{g}$ (\tikzcircle[black,fill=blue_light]{15pt}) is the
projection of $\boldsymbol{g}_{term}$ onto the sphere $\mathcal{S}$.}\tabularnewline
\hline 
$\mathbf{p}_{i}(t)$ & Predicted trajectory of an obstacle $i$\add{, and committed trajectory of agent $i$.}\tabularnewline
\hline 
$\mathcal{T}_{j}$\add{, $\gamma_j$} & \add{$\mathcal{T}_{j}$ is a } uniform discretization of $\ensuremath{[t_{p+j},t_{p+j+1}]}$ (timespan
of interval $j$ of the trajectory of agent $s$) with step size $\gamma_{j}$
and such that $t_{p+j},t_{p+j+1}\in\text{\ensuremath{\mathcal{T}_{j}}}$.\tabularnewline
\hline 
$\mathbf{p}_{i}\left(\mathcal{T}_{j}\right)$ & $\left\{ \mathbf{p}_{i}(t)\;|\;t\in\mathcal{T}_{j}\right\} $ \tabularnewline
\hline 
\end{tabular}
\par\end{centering}
}
\end{table}

This paper will use the notation shown in Table \ref{tab:Notation-used-in},
together with the following two definitions:

\begin{itemize}
\item \textbf{Agent: }Element of the environment with the ability to exchange
information and take decisions accordingly (i.e., an agent can change
its trajectory given the information received from the environment). 
\item \textbf{Obstacle: }Element of the environment that moves on its own
without consideration of the trajectories of other elements in the
environment. An obstacle can be static or dynamic. 
\end{itemize}
Note that here we are calling dynamic obstacles what some works in
the literature call \emph{non-cooperative agents}. 

This paper will also use clamped uniform B-Splines, which are B-Splines
defined by $n+1$ control points $\{\boldsymbol{q}_{0},\ldots,\boldsymbol{q}_{n}\}$
and $m+1$ knots $\{t_{0},t_{1},\ldots,t_{m}\}$ that satisfy:
\[
\underbrace{t_{0}=...=t_{p}}_{p+1\;\text{knots}}<\underbrace{{\color{black}t_{p+1}<...<t_{m-p-1}}}_{\text{Internal Knots}}<\underbrace{t_{m-p}=...=t_{m}}_{p+1\;\text{knots}}
\]
and where the internal knots are equally spaced by $\Delta t$ (i.e.
$\Delta t:=t_{k+1}-t_{k}\;\;\forall k=\{p,\ldots,m-p-1\}$). The relationship $m=n+p+1$ holds, and there are
in total $m-2p=n-p+1$ intervals. Each interval $j\in J$ is defined in $t\in\left[t_{p+j},t_{p+j+1}\right]$.
In this paper we will use $p=3$ (i.e. cubic B-Splines). Hence, each interval will be a polynomial of degree $3$, and it is guaranteed
to lie within the convex hull of its $4$ control points $\{\boldsymbol{q}_{j}$,
$\boldsymbol{q}_{j+1}$, $\boldsymbol{q}_{j+2}$, $\boldsymbol{q}_{j+3}\}$.
Moreover, clamped B-Splines are guaranteed to pass through the first
and last control points ($\boldsymbol{q}_{0}$ and $\boldsymbol{q}_{n}$). The velocity and acceleration of a B-Spline are B-Splines of degrees $p-1$ and $p-2$ respectively, whose control points are given by (\cite{zhou2019robust}):
\begin{align*} 
\boldsymbol{v}_{l}&=\frac{p\left(\boldsymbol{q}_{l+1}-\boldsymbol{q}_{l}\right)}{t_{l+p+1}-t_{l+1}}\quad\forall l\in L\backslash\{n\} \\ 
\boldsymbol{a}_{l}&=\frac{\left(p-1\right)\left(\boldsymbol{v}_{l+1}-\boldsymbol{v}_{l}\right)}{t_{l+p+1}-t_{l+2}}\quad\forall l\in L\backslash\{n-1,n\}
\end{align*}

\add{Finally, and as shown in Table \ref{tab:Notation-used-in}, to obtain $\boldsymbol{g}$ we project the terminal goal $\boldsymbol{g}_\text{term}$ to a sphere $\mathcal{S}$ centered on $\boldsymbol{d}$. This is done only for simplicity, and other possible way would be to choose $\boldsymbol{g}$ as the intersection between $\mathcal{S}$ and a piecewise linear path that goes from $\boldsymbol{d}$ to $\boldsymbol{g}_\text{term}$, and that avoids the static obstacles (and potentially the dynamic obstacles or agents as well). If a voxel grid of the environment is available, this piecewise linear path could be obtained by running a search-based algorithm, as done in \cite{tordesillas2019faster}.  }

\newpage
\section{Assumptions \label{sec:Assumptions}}

This paper relies on the following \add{four} assumptions:
\begin{itemize}
    \item Let $\mathbf{p}^{\text{real}}_{i}(t)$ denote the real future trajectory of
an obstacle $i$, and $\mathbf{p}_{i}(t)$ the one obtained
by a given tracking and prediction algorithm. The smallest dimensions
of the axis-aligned box $D_{ij}$ for which 
\[
\mathbf{p}^{\text{real}}_{i}(t)\in\text{conv}\left(D_{ij}\oplus\mathbf{p}_i(\mathcal{T}_{j})\right)\quad\forall t\in[t_{p+j},t_{p+j+1}]
\]

is satisfied will be denoted as $2\left(\boldsymbol{\alpha}_{ij}+\boldsymbol{\beta}_{ij}\right)\in\mathbb{R}^{3}$.
\add{Here, $\mathcal{T}_{j}$ is a uniform discretization of $\ensuremath{[t_{p+j},t_{p+j+1}]}$ with step size $\gamma_j$ (see Table \ref{tab:Notation-used-in}),} $\boldsymbol{\alpha}_{ij}$ represents the error associated with
the prediction and $\boldsymbol{\beta}_{ij}$ the one associated with
the discretization of the trajectory of the obstacle. The values $\boldsymbol{\alpha}_{ij},\boldsymbol{\beta}_{ij}$
and $\gamma_{j}$ are assumed known. This assumption is needed to be able to obtain an outer polyhedral approximation of the Minkowski sum of a bounding box and any continuous trajectory of an obstacle (Sec. \ref{subsec:polyhedral_rep_obstacles}).

\item Similar to other works in the literature (see \cite{liu2018towards} for instance), we assume that an agent
 can communicate without delay with other agents. \add{Specifically, we assume that the planning agent has access to the committed trajectory $\mathbf{p}_{i}(t)$ of agent $i$ when this condition holds:
 $$
 \exists t\in[t_{\text{in}},t_{\text{f}}]\;\;\text{s.t.}\;\;\left(\mathbf{p}_{i}(t)\oplus B_{i}\right)\cap\left(\mathcal{S}\oplus B_{s}\right)\neq\emptyset
 $$
This condition ensures that the agent $s$ knows the trajectories of the agents whose committed trajectories, inflated with their AABBs, pass through the sphere $\mathcal{S}$ (inflated with $B_s$) during the interval $[t_{\text{in}},t_{\text{f}}]$.} Note also that all the agents have the same reference time, but trigger the planning iterations asynchronously.
\item Two agents do not commit to a new trajectory at the very same time. Note that, as time is continuous, the probability of this assumption not being true is essentially zero. \add{Letting $t_4$ denote the time when a UAV commits to a trajectory, the} reason behind this assumption is to guarantee that it is safe for a UAV to commit to a trajectory at $t=t_4$ having checked all the committed trajectories of other agents at $t<t_4$ (this will be explained in detail in Sec. \ref{sec:deconfliction}). 
\add{
\item Finally, we assume for simplicity that the obstacles do not rotate (and hence $B_i$ is constant for an obstacle~$i$). However, this is not a fundamental assumption in MADER: To take into account the rotation of the objects, one could still use MADER, but use for the inflation (Sec.~\ref{sec:PolyhedralRepresentations}) the largest AABB that contains all the rotations of the obstacle during a specific interval~$j$.}

\end{itemize}

\section{Polyhedral representations}\label{sec:PolyhedralRepresentations}

To avoid the computational burden of imposing infinitely-many constrains
to separate two trajectories, we need to compute a tight polyhedral
outer representation of every interval of the optimized trajectory
(trajectory that agent $s$ is trying to obtain), the trajectory of
the other agents and the trajectory of other obstacles (see also Table \ref{tab:Polyhedral-Representations-of}). 
\begin{table}
\begin{centering}
\caption{Polyhedral representations of interval $j$ from the point of view
of agent $s$. Here, $\mathcal{R}_{ij}^{\text{MV}}$ denotes the set
of MINVO control points of every interval of the trajectory of agent
$i$ that falls in $[t_{\text{in}}+j\Delta t,t_{\text{in}}+(j+1)\Delta t]$
(timespan of the interval $j$ of the trajectory of agent $s$). \label{tab:Polyhedral-Representations-of}}
\par\end{centering}
\noindent\resizebox{\columnwidth}{!}{%
\begin{centering}
\begin{tabular}{|>{\centering}p{2.3cm}|>{\centering}m{1.3cm}|>{\centering}m{3.2cm}|>{\centering}m{2.2cm}|}
\cline{2-4} \cline{3-4} \cline{4-4} 
\multicolumn{1}{>{\centering}p{2.3cm}|}{} & \textbf{Trajectory} & \textbf{Inflation} & \textbf{Polyhedral Repr.}\tabularnewline
\hline 
\textbf{Agent $s$} & B-Spline & No Inflation & \add{$\text{conv}\left(\mathcal{Q}_{j}^{\text{MV}}\right)$}\tabularnewline
\hline 
\textbf{Other agents $i\in I$} & B-Spline & $B_{i}'=B_{i}$ inflated with $\boldsymbol{\eta}_{s}$ & $\text{conv}\left(B_{i}'\oplus\mathcal{R}_{ij}^{\text{MV}}\right)$\tabularnewline
\hline 
\textbf{Obstacles $i\in I$} & Any  & $B_{i}'=B_{i}$ inflated with $\boldsymbol{\eta}_{s}+2\left(\boldsymbol{\beta}_{ij}+\boldsymbol{\alpha}_{ij}\right)$ & $\text{conv}\left(B_{i}'\oplus\boldsymbol{p}_{i}\left(\mathcal{T}_{j}\right)\right)$\tabularnewline
\hline 
\end{tabular}
\par\end{centering}
}
\vskip -0.4cm
\end{table}

\subsection{Polyhedral Representation of the trajectory of the agent~$s$}\label{sec:polyhedral_representation}

\begin{figure}
\begin{centering}
\includegraphics[width=1\columnwidth]{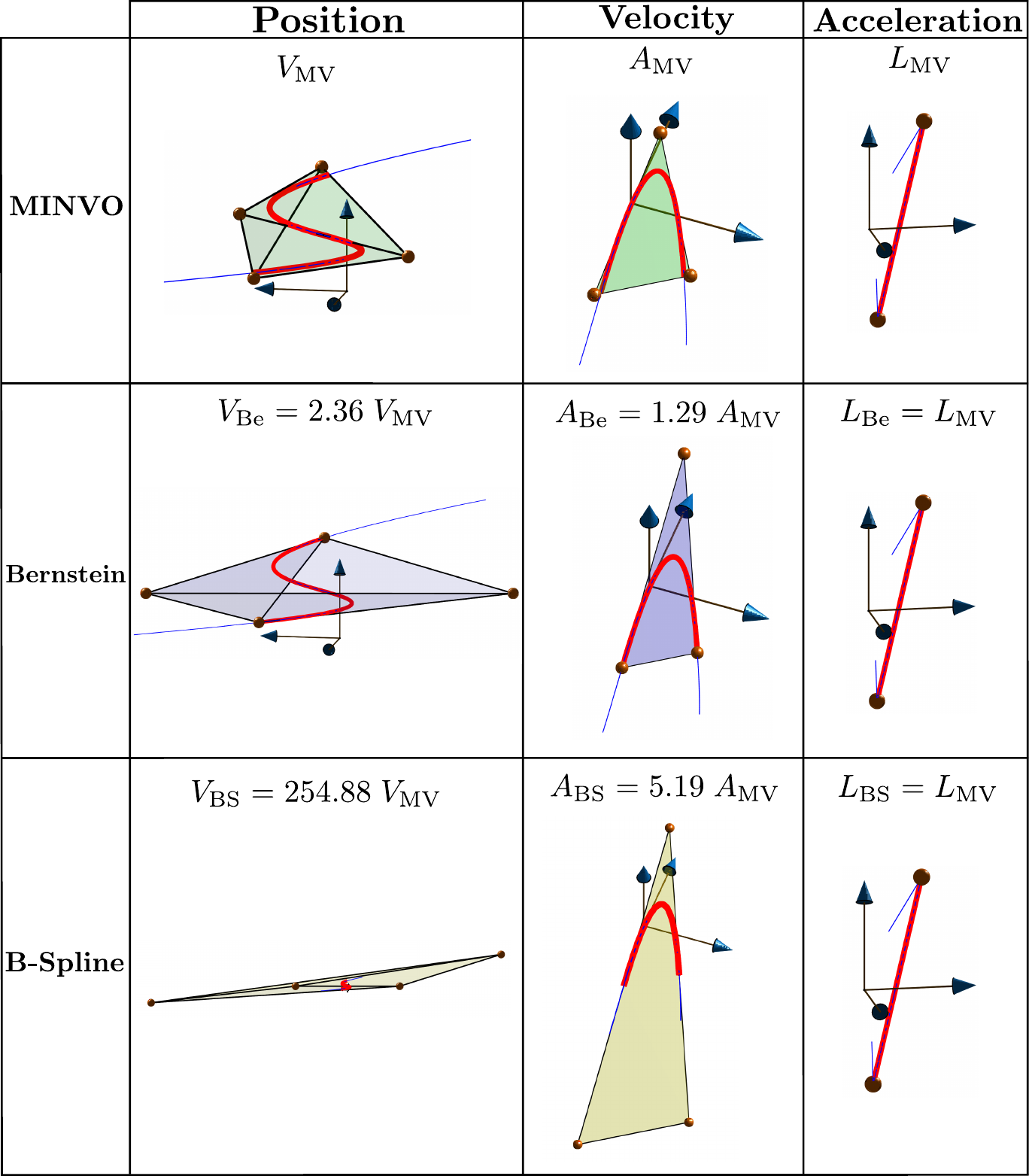}
\par\end{centering}
\caption{Comparison of the volumes, areas and lengths obtained by the MINVO
basis (ours), Bernstein basis (used by the Bézier curves) and B-Spline
basis for an interval (\textcolor{red}{\rule[0.05cm]{0.7cm}{0.6mm}})
of a given uniform B-Spline (\textcolor{blue}{\rule[0.06cm]{0.7cm}{0.6mm}}).
In the acceleration space, the three bases generate the same control
points. \label{fig:comparisonBsBezierMinvoPosVelAccel} }
\end{figure}

When using B-Splines, one common way to obtain an outer
polyhedral representation for each interval is to use the polyhedron
defined by the control points of each interval. As the functions in the B-Spline basis are positive and form a partition of unity, this polyhedron is guaranteed
to completely contain the interval. However, this approximation is
far from being tight, leading therefore to great conservatism both
in the position and in the velocity space. To mitigate this, \cite{tang2019real} used the Bernstein basis for the constraints
in the velocity space. Although this basis generates a polyhedron
smaller than the B-Spline basis, it is still conservative, as this
basis does not minimize the volume of this polyhedron. We instead
use both in position \emph{and} velocity
space our recently derived MINVO basis \cite{tordesillas2020minvo}
that, by construction, is a polynomial basis that attempts to obtain
the simplex with minimum volume that encloses a given polynomial curve.
As shown in Fig. \ref{fig:comparisonBsBezierMinvoPosVelAccel}, this
basis achieves a volume that is 2.36 and 254.9 times smaller (in the
position space) and $1.29$ and $5.19$ times smaller (in the velocity
space) than the Bernstein and B-Spline bases respectively. For each
interval $j$, the vertexes of the MINVO control points ($\boldsymbol{Q}_{j}^{\text{MV}}$ and $\boldsymbol{V}_{j}^{\text{MV}}$ for position and velocity respectively)
and the B-Spline control points ($\boldsymbol{Q}_{j}^{\text{BS}}$, $\boldsymbol{V}_{j}^{\text{BS}}$) are related
as follows:
\add{\begin{align} \label{eq:BS2MV}
\begin{split}
\boldsymbol{Q}_{j}^{\text{MV}}&=\boldsymbol{Q}_{j}^{\text{BS}}\boldsymbol{A}_{\text{pos}}^{\text{BS}}(j)\left(\boldsymbol{A}_{\text{pos}}^{\text{MV}}\right)^{-1} \\
\boldsymbol{V}_{j}^{\text{MV}}&=\boldsymbol{V}_{j}^{\text{BS}}\boldsymbol{A}_{\text{vel}}^{\text{BS}}(j)\left(\boldsymbol{A}_{\text{vel}}^{\text{MV}}\right)^{-1}
\end{split}
\end{align}}
where the matrices $\boldsymbol{A}$ are known, and are available
in our recent work \cite{tordesillas2020minvo} (for the MINVO basis)
and in \cite{qin2000general} (for the Bernstein and B-Spline bases). \add{For the B-Spline bases, and because we are using clamped uniform splines, the matrices $\boldsymbol{A}_{\text{pos}}^{\text{BS}}(j)$ and $\boldsymbol{A}_{\text{vel}}^{\text{BS}}(j)$ depend on the interval~$j$. Eq.~\ref{eq:BS2MV}, together with the fact that $\boldsymbol{V}_{j}^{\text{BS}}$ is a linear combination of $\boldsymbol{Q}_{j}^{\text{BS}}$, allow us to write }
	\add{\begin{align} \label{eq:BS2MV_sets}
			\begin{split}
				\mathcal{Q}_{j}^{\text{MV}}&=\ffQ \\
				\mathcal{V}_{j}^{\text{MV}}&=\hhQ
			\end{split}
	\end{align}}
\add{where $\ff{\cdot}$ and $\hh{\cdot}$ are known linear functions.}

\begin{figure}
	\begin{centering}
		\includegraphics[width=1\columnwidth]{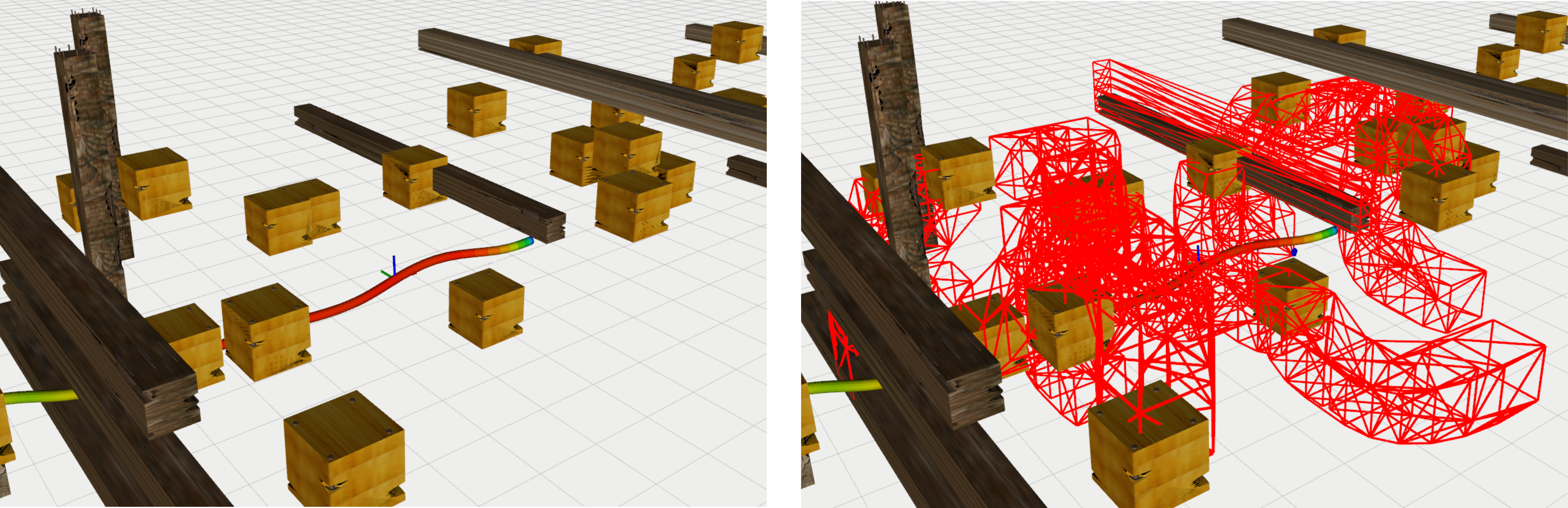}
		\par\end{centering}
	\caption{To impose collision-free constraints, MADER uses polyhedral representations of each interval of the trajectories of other agents/obstacles. On the left a given scenario with dynamic obstacles and on the right the polyhedral representations obtained (in red).\label{fig:intervals_rviz}}	
\end{figure}

\begin{figure*}
\begin{centering}
\modifiedFig{\includegraphics[width=1\textwidth]{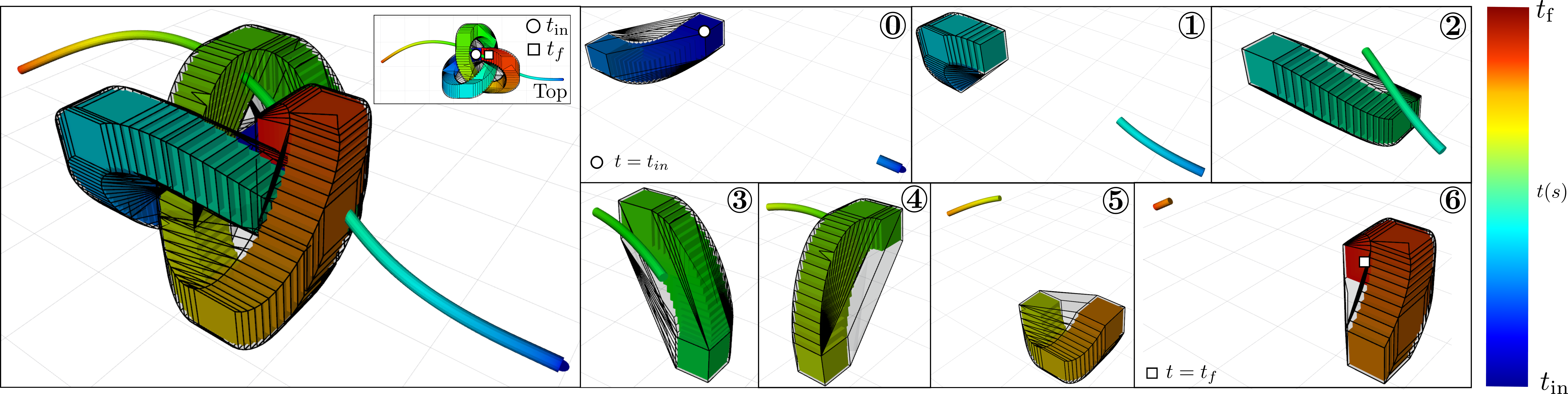}}
\par\end{centering}
\caption{Example of a trajectory avoiding a dynamic obstacle. The obstacle has a box-like shape and is moving following a trefoil knot trajectory. The trajectory of the obstacle is divided into as many segments as the optimized trajectory has. An outer polyhedral representation (whose edges are shown as black lines) is computed for each of theses segments, and each segment of the trajectory avoids these polyhedra.\label{fig:init_guess}}
\end{figure*}

\begin{figure*}
\begin{centering}
\modifiedFig{\includegraphics[width=1\textwidth]{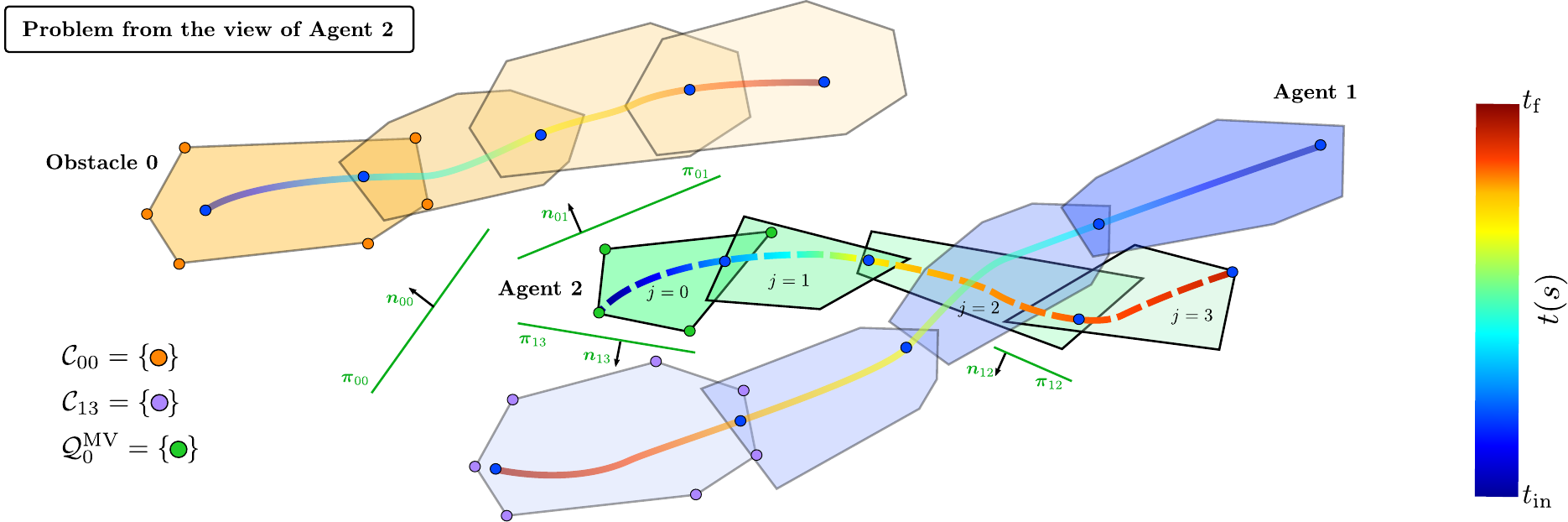}}
\par\end{centering}
\caption{Collision-free constraints between agent 2 and both the obstacle 0
and agent 1. This figure is from the view of agent 2. $t_{\text{in}}$
and $t_{\text{f}}$ are the initial and final times of the trajectory
being optimized (dashed lines), and they are completely independent of the initial
and final optimization times of agent 1. $\mathcal{Q}_{0}^{\text{MV}}$
are the control points of the interval $0$ of the optimized trajectory using the
MINVO basis. $\mathcal{C}_{13}$ are the vertexes of the convex hull
of the vertexes of the control points of all the intervals of the
trajectory of agent 1 that fall in $[t_{\text{f}}-\Delta t,t_{\text{f}}]$.
Note that the trajectories and polyhedra are in 3D, but they are represented
in 2D for visualization purposes. \label{fig:explanation_trajs}}
\end{figure*}

\subsection{Polyhedral Representation of the trajectory of other agents}

We first increase the sides of $B_{i}$ by $\boldsymbol{\eta}_{s}$
to obtain the inflated box $B_{i}'$. Now, note that the trajectory
of the agent $i\neq s$ is also a B-Spline, but its initial and final
times can be different from $t_{\text{in}}$ and $t_{\text{f}}$
(initial and final times of the trajectory that agent $s$ is optimizing).
Therefore, to obtain the polyhedral representation of the trajectory
of the agent $i$ in the intervals $[t_{\text{in}},t_{\text{in}}+\Delta t]$, $[t_{\text{in}}+\Delta t,t_{\text{in}}+2\Delta t]$, \ldots, $[t_{\text{f}}-\Delta t,t_{\text{f}}]$
we first compute the MINVO control points of every interval of the
trajectory of agent $i$ that falls in one of these intervals. The
convex hull of the boxes $B_{i}'$ placed in every one of these control
points will be the polyhedral representation of the interval $j$ of the trajectory of the agent $i$. We denote the vertexes of this outer polyhedral representation as $\mathcal{C}_{ij}$.

\subsection{Polyhedral Representation of the trajectory of the obstacles} \label{subsec:polyhedral_rep_obstacles}

For each interval $j$ we first increase the sides of $B_{i}$ by
$\boldsymbol{\eta}_{s}+2(\boldsymbol{\beta}_{ij}+\boldsymbol{\alpha}_{ij})$, and denote this
inflated box $B_{i}'$. Here $\boldsymbol{\beta}_{ij}$ and $\boldsymbol{\alpha}_{ij}$ are
the values defined in Sec. \ref{sec:Assumptions}.  
We then place $B_{i}'$ in $\mathbf{p}_i(\mathcal{T}_j)$, where $\mathbf{p}_i(\mathcal{T}_j)$ denotes the set of positions of the obstacle~$i$ at the times $\mathcal{T}_j$ (see Table \ref{tab:Notation-used-in}) and compute the convex hull of all the vertexes of these boxes.
Given the first assumption of Sec. \ref{sec:Assumptions},
this guarantees that the convex hull obtained is an outer approximation of all the 3D space occupied by the obstacle~$i$ (inflated by the size of the agent~$s$) during the interval~$j$. 
The static obstacles are treated in the same way, with $\mathbf{p}_i(t)=\text{constant}$. An example of these polyhedral representations 
is shown in Fig.~\ref{fig:intervals_rviz}.

\section{Optimization And Initial Guess}

\subsection{Collision-free constraints}\label{subsec:collisionfreeconst}

Once the polyhedral approximation of the trajectories of the other
obstacles/agents have been obtained, we enforce the collision-free
constraints between these polyhedra and the ones of the optimized
trajectory as follows: we introduce the planes $\boldsymbol{\pi}_{ij}$
(characterized by $\boldsymbol{n}_{ij}$ and $d_{ij}$) that separate
them as decision variables in the optimization problem and force this way the separation between the vertexes in $\mathcal{C}_{ij}$ and the MINVO control points $\mathcal{Q}_{j}^{\text{MV}}$  (see Figs. \ref{fig:init_guess} and
\ref{fig:explanation_trajs}):
\begin{equation}
\begin{aligned} \label{eq:collision-free}
\boldsymbol{n}_{ij}^{T}\boldsymbol{c}+d_{ij}>0\quad\forall\boldsymbol{c}\in\mathcal{C}_{ij},\;\forall i\in I,j\in J\\
\boldsymbol{n}_{ij}^{T}\boldsymbol{q}+d_{ij}<0\quad\forall\boldsymbol{q}\in\mathcal{Q}_{j}^{\text{MV}},\;\forall j\in J
\end{aligned}
\end{equation}

\subsection{Other constraints}\label{sec:OtherConstraints}

The initial condition (position, velocity and acceleration) is imposed by $\mathbf{x}(t_{\text{in}})=\mathbf{x}_{in} $. Note that $\boldsymbol{p}_{\text{in}}$, $\boldsymbol{v}_{\text{in}}$
and $\boldsymbol{a}_{\text{in}}$ completely determine $\boldsymbol{q}_{0}$,
$\boldsymbol{q}_{1}$ and $\boldsymbol{q}_{2}$, so these control
points are not included as decision variables.

For the final condition, we use a final stop condition imposing the constraints $\mathbf{v}(t_{\text{f}})=\mathbf{v}_{f}=\boldsymbol{0}$ and $\mathbf{a}(t_{\text{f}})=\mathbf{a}_{f}=\boldsymbol{0}$. These conditions require $\boldsymbol{q}_{n-2}=\boldsymbol{q}_{n-1}=\boldsymbol{q}_{n}$, so the control points $\boldsymbol{q}_{n-1}$ and $\boldsymbol{q}_{n}$ can also be excluded from the set of decision variables.  The final position is included as a penalty cost $\left\Vert \boldsymbol{q}_{n-2}-\boldsymbol{g}\right\Vert ^{2} _2$ in the objective function, weighted with a parameter $\omega\ge0$. Here $\boldsymbol{g}$ is the goal (projection of the $\boldsymbol{g}_{\text{term}}$
onto a sphere $\mathcal{S}$ of radius $r$ around $\boldsymbol{d}$,
see Table \ref{tab:Notation-used-in}). Note that, as we are using clamped uniform B-Splines with a final stop condition, $\boldsymbol{q}_{n-2}$ coincides with the last position of the B-Spline. \add{The reason of adding this penalty cost for the final position, instead of including $\boldsymbol{q}_{n-2}=\boldsymbol{g}$ as a hard constraint, is that a hard constraint can easily lead to infeasibility if the heuristics used for the total time $(t_{\text{f}}-t_{\text{in}})$ underestimates the time needed to reach~$\boldsymbol{g}$.}

To force the trajectory generated to be inside the sphere $\mathcal{S}$, we impose the constraint 
\begin{equation}\label{eq:sphere_constraints}
	\left\Vert \boldsymbol{q}-\boldsymbol{d}\right\Vert _{2}^{2}\le r^{2}\qquad\forall\boldsymbol{q}\in\mathcal{Q}_{j}^{\text{MV}},\;\forall j\in J
\end{equation}
Moreover, we also add the constraints on the maximum velocity and acceleration:
\begin{equation}
	\begin{aligned}\label{eq:max_vel_accel}
&\qquad \text{abs}\left(\boldsymbol{v}\right) \le \boldsymbol{v}_{\text{max}}\quad\forall\boldsymbol{v}\in\mathcal{V}_{j}^{\text{MV}},\;\forall j\in J & \\
&\qquad \text{abs}\left(\boldsymbol{a}_{l}\right) \le \boldsymbol{a}_{\text{max}}\quad  \forall l\in L\backslash\{n-1,n\}
  \end{aligned}
\end{equation}

where we are using the MINVO velocity control points for the velocity constraint. For the acceleration constraint, the B-Spline and MINVO control points are the same (see Fig. \ref{fig:comparisonBsBezierMinvoPosVelAccel}). \add{Note that the velocity and acceleration are constrained independently on each one of the axes $\{x,y,z\}$}.

\subsection{Control effort}\label{subsec:controleffort}

The evaluation of a cubic clamped uniform B-Spline in an interval
$j\in J$ can be done as follows \cite{qin2000general}:
\[
\mathbf{p}(t)=\boldsymbol{Q}_{j}^{\text{BS}}\boldsymbol{A}_{\text{pos}}^{\text{BS}}(j)\underbrace{\left[\begin{array}{c}
u_{j}^{3}\\
u_{j}^{2}\\
u_{j}\\
1
\end{array}\right]}_{:=\boldsymbol{u}_{j}}
\]
\add{where $u_{j}:=\frac{t-t_{p+j}}{t_{p+j+1}-t_{p+j}}$, }
$t\in [t_{p+j},t_{p+j+1}]$ and $\boldsymbol{A}_{\text{pos}}^{\text{BS}}(j)$ is a known matrix that depends on each interval. Specifically, and
with the knots chosen, we will have $\boldsymbol{A}_{\text{pos}}^{\text{BS}}(0)\neq\boldsymbol{A}_{\text{pos}}^{\text{BS}}(1)\neq\boldsymbol{A}_{\text{pos}}^{\text{BS}}(2)=...=\boldsymbol{A}_{\text{pos}}^{\text{BS}}(m-2p-3)\neq\boldsymbol{A}_{\text{pos}}^{\text{BS}}(m-2p-2)\neq\boldsymbol{A}_{\text{pos}}^{\text{BS}}(m-2p-1)$.
Now, note that
\[
\frac{d^{r}\mathbf{p}(t)}{dt^{r}}=\frac{1}{\Delta t^{r}}\boldsymbol{Q}_{j}^{\text{BS}}\boldsymbol{A}_{\text{pos}}^{\text{BS}}(j)\frac{d^{r}\boldsymbol{u}_{j}}{{du_j}^{r}}
\]
Therefore, as the jerk is constant in each interval (since
$p=3$), the control effort is:
\begin{equation}\label{eq:control_effort}
\int_{t_{\text{in}}}^{t_{\text{f}}}\left\Vert \mathbf{j}(t)\right\Vert ^{2}dt\propto\sum_{j\in J}\left\Vert \boldsymbol{Q}_{j}^{\text{BS}}\boldsymbol{A}_{\text{pos}}^{\text{BS}}(j)\left[\begin{array}{c}
6\\
0\\
0\\
0
\end{array}\right]\right\Vert ^{2} _2
\end{equation}

\subsection{Optimization Problem}\label{sec:opt_problem}

Given the constraints and the objective function explained above, the optimization problem solved is as follows\footnote{\add{In the optimization problem, $\forall i$ and $\forall j$ denote, respectively, $\forall i\in I$ and $\forall j\in J$.}}: 

\begin{empheq}[box=\fbox]{flalign*}
&\underset{\mathcal{Q}_{j}^{\text{BS}},\boldsymbol{n}_{ij},d_{ij}}{\boldsymbol{\min}}\sum_{j\in J}\left\Vert \boldsymbol{Q}_{j}^{\text{BS}}\boldsymbol{A}_{\text{pos}}^{\text{BS}}(j)\left[\begin{array}{c} 6\\ 0\\ 0\\ 0 \end{array}\right]\right\Vert ^{2} _2+\omega\left\Vert \boldsymbol{q}_{n-2}-\boldsymbol{g}\right\Vert ^{2} _2 & \\
&\text{s.t.} & \\ \quad
   &\quad\mathbf{x}(t_{\text{in}})=\mathbf{x}_{in} & \\
   &\quad\mathbf{v}(t_{\text{f}})=\mathbf{v}_{f}=\boldsymbol{0} & \\
   &\quad\mathbf{a}(t_{\text{f}})=\mathbf{a}_{f}=\boldsymbol{0} & \\
   &\quad\boldsymbol{n}_{ij}^{T}\boldsymbol{c}+d_{ij}>0\quad\forall\boldsymbol{c}\in\mathcal{C}_{ij},\add{\;\forall i,j} & \\
&\quad\boldsymbol{n}_{ij}^{T}\boldsymbol{q}+d_{ij}<0\quad\forall\boldsymbol{q}\in\mathcal{Q}_{j}^{\text{MV}}\add{:=\ffQ},\add{\;\forall i,j} & \\
&\quad\left\Vert \boldsymbol{q}-\boldsymbol{d}\right\Vert _{2}^{2}\le r^{2}\quad\forall\boldsymbol{q}\in\mathcal{Q}_{j}^{\text{MV}}\add{:=\ffQ,\;\forall j} & \\
&\quad \text{abs}\left(\boldsymbol{v}\right) \le \boldsymbol{v}_{\text{max}}\quad\forall\boldsymbol{v}\in\mathcal{V}_{j}^{\text{MV}}\add{:=\hhQ},\add{\;\forall j} & \\
&\quad \text{abs}\left(\boldsymbol{a}_{l}\right) \le \boldsymbol{a}_{\text{max}}\quad  \forall l\in L\backslash\{n-1,n\}
\end{empheq}
\vspace{0.2cm}

This problem is clearly nonconvex
since we are minimizing over the control points \emph{and }the planes
$\boldsymbol{\pi}_{ij}$ (characterized by $\boldsymbol{n}_{ij}$ and
$d_{ij}$). 
\add{
Note also that the decision variables are the B-Spline control points $\mathcal{Q}_{j}^{\text{BS}}$. In the constraints, the MINVO control points $\mathcal{Q}_{j}^{\text{MV}}$ and $\mathcal{V}_{j}^{\text{MV}}$ are simply linear transformations 
 of the decision variables (see Eq.~\ref{eq:BS2MV_sets}).}
We solve this problem using the augmented Lagrangian method \cite{conn1991globally,birgin2008improving},
and with the globally-convergent method-of-moving-asymptotes (MMA)
\cite{svanberg2002class} as the subsidiary optimization algorithm. The interface used for these algorithms is NLopt \cite{nlopt20}.
The time allocated per trajectory is chosen before the optimization
as $(t_{\text{f}}-t_{\text{in}})=\frac{\left\Vert \boldsymbol{g}-\boldsymbol{d}\right\Vert _{2}}{v_{max}}$.

\subsection{Initial Guess}\label{sec:InitialGuess}

\begin{algorithm2e}[t]
	\footnotesize
	\DontPrintSemicolon
	
	\SetKwRepeat{Do}{do}{while}
	\SetKwFunction{FMain}{\textbf{GetInitialGuess}}
	\SetKwProg{Pn}{Function}{:}{\KwRet}
	\Pn{\FMain{}}{
		
		Compute $\boldsymbol{q}_{0}$, $\boldsymbol{q}_{1}$ and $\boldsymbol{q}_{2}$
		from $\boldsymbol{p}_{\text{in}}$, $\boldsymbol{v}_{\text{in}}$
		and $\boldsymbol{a}_{\text{in}}$\label{line:computeq0q1q2}
		
		Add $\boldsymbol{q}_{2}$ to $Q$\label{line:addq2}
		
		\While{$Q$ is not empty}{
			
			$\boldsymbol{q}_{l}\leftarrow$First element of $Q$\label{line:getFirstElem1}
			
			Remove first element of $Q$\label{line:getFirstElem2}
			
			$\mathcal{M}\leftarrow$Uniformly sample $\boldsymbol{v}_{l}$ satisfying
			$v_{max}$ and $a_{max}$
			
			\If{any of the conditions \ref{enu:collision1}-\ref{enu:sizeOfV}
				is true}{
				
				\textbf{continue}}
			
			\If{$\left\Vert \boldsymbol{q}_{l}-\boldsymbol{g}\right\Vert _{2}<\epsilon''$
				and $l=(n-2)$}{\label{line:checkInGoal1}
				
				$\boldsymbol{q}_{n-1}\leftarrow\boldsymbol{q}_{n-2}$
				
				$\boldsymbol{q}_{n}\leftarrow\boldsymbol{q}_{n-2}$
				
				\Return{ $\{\boldsymbol{q}_{0},\boldsymbol{q}_{1},\boldsymbol{q}_{2},\ldots,\boldsymbol{q}_{n-2},\boldsymbol{q}_{n-1},\boldsymbol{q}_{n}\}\cup\boldsymbol{\pi}_{ij}$}
				
				\label{line:checkInGoal2}}
			
			\For{every $\boldsymbol{v}_{l}$ in $\mathcal{M}$}{\label{line:Add_samples_toQ1}
				
				$\boldsymbol{q}_{l+1}\leftarrow\boldsymbol{q}_{l}+\frac{t_{l+p+1}-t_{l+1}}{p}\boldsymbol{v}_{l}$
				
				Store in $\boldsymbol{q}_{l+1}$ a pointer to $\boldsymbol{q}_{l}$\label{line:Store-in-Q}
				
				Add $\boldsymbol{q}_{l+1}$ to $Q$\label{line:Add-to-Q}
				
				\label{line:Add_samples_toQ-2}}
			
		}
		
		\Return{Closest Path found}\label{line:return}
		
	}
	\normalsize
	\caption{Octopus Search \label{IR}}
	\label{algo: octopus}
\end{algorithm2e}

To obtain an initial guess (which consists of both the control points
$\{\boldsymbol{q}_{0},\ldots,\boldsymbol{q}_{n}\}^{\text{BS}}$ and the planes
$\boldsymbol{\pi}_{ij}$), we use the Octopus Search
algorithm shown in Alg. \ref{algo: octopus}. The Octopus Search takes inspiration from A\textsuperscript{*} \cite{hart1968formal},
but it is designed to work with B-Splines, handle dynamic obstacles/agents,
and use the MINVO basis for the collision check. Each control point
will be a node in the search. All the open nodes are kept in a priority
queue $Q$, in which the elements are ordered in increasing order
of $f=g+\epsilon h$, where $g$ is the sum of the distances (between successive 
control points) from $\boldsymbol{q}_0$ to the current node \add{(cost-to-come)}, $h$ is the distance from the
current node to the goal \add{(heuristics of the cost-to-go)}, and $\epsilon$ is the bias. \add{Similar to A\textsuperscript{*}, this ordering of the priority queue makes nodes with lower $f$ be explored first.}

The way the algorithm works is as follows: First, we compute the control
points ${\boldsymbol{q}_{0},\;\boldsymbol{q}_{1},\;\boldsymbol{q}_{2}}$,
which are determined from $\boldsymbol{p}_{\text{in}}, \boldsymbol{v}_{\text{in}}$
and $\boldsymbol{a}_{\text{in}}$. After adding $\boldsymbol{q}_{2}$
to the queue $Q$ (line \ref{line:addq2}), we run the following loop
until there are no elements in $Q$: First we store in $\boldsymbol{q}_{l}$
the first element of $Q$, and remove it from $Q$ (lines \ref{line:getFirstElem1}-\ref{line:getFirstElem2}).
Then, we store in a set $\mathcal{{M}}$ velocity samples for $\boldsymbol{v}_{l}$
that satisfy both $v_{\text{max}}$ and $a_{\text{max}}$\footnote{In these velocity samples, we use the B-Spline velocity control points, to avoid the dependency with past velocity control points that appears when using the MINVO or Bernstein bases. But note that this is only for the initial guess, in the optimization problem the MINVO velocity control points are used.}.  After this,
we discard the current $\boldsymbol{q}_{l}$ if any of these conditions
are true (l.s. denotes linearly separable):

\vspace{0.1cm}

\begin{enumerate}
\item \label{enu:collision1}$\mathcal{Q}_{l-3}^{\text{MV}}$ is not l.s. from
$\mathcal{C}_{i,l-3}$ for some $i\in I$. 
\item \label{enu:collision2}$l=(n-2)$ and $\mathcal{Q}_{n-4}^{\text{MV}}$ is
not l.s. from $\mathcal{C}_{i,n-4}$ for some $i\in I$.
\item \label{enu:collision3}$l=(n-2)$ and $\mathcal{Q}_{n-3}^{\text{MV}}$ is
not l.s. from $\mathcal{C}_{i,n-3}$ for some $i\in I$.
\item \label{enu:sphere}$\left\Vert \boldsymbol{q}_{l}-\boldsymbol{d}\right\Vert _{2}>r$.
\item \label{enu:voxel}$\left\Vert \boldsymbol{q}_{l}-\boldsymbol{q}_{k}\right\Vert _{\infty}\le\epsilon'$
for some $\boldsymbol{q}_{k}$ already added to $Q$. 
\item \label{enu:sizeOfV}Cardinality of $\mathcal{{M}}$ is zero.
\end{enumerate}
\vspace{0.1cm}

Condition \ref{enu:collision1} ensures that the convex hull of
$\mathcal{Q}_{l-3}^{\text{MV}}$ does not collide with any interval $l-3$
of other obstacle/agent $i\in I$. The linear separability is checked by
solving the following feasibility linear problem for the interval
$j=l-3$ of every obstacle/agent $i\in I$:
\begin{equation}
\label{eqn:LP_Problem}
\begin{array}{cc}
\boldsymbol{n}_{ij}^{T}\boldsymbol{c}+d_{ij}>0 & \forall\boldsymbol{c}\in\mathcal{C}_{ij}\\
\boldsymbol{n}_{ij}^{T}\boldsymbol{q}+d_{ij}<0 & \forall\boldsymbol{q}\in\mathcal{Q}_{j}^{\text{MV}}
\end{array}
\end{equation}
where the decision variables are the planes $\boldsymbol{\pi}_{ij}$
(defined by $\boldsymbol{n}_{ij}$ and $d_{ij}$). We solve this problem
using GLPK \cite{glpkSolver20}. Note that we also need to check the
conditions \ref{enu:collision2} and \ref{enu:collision3} due to
the fact that $\boldsymbol{q}_{n-2}=\boldsymbol{q}_{n-1}=\boldsymbol{q}_{n}$ and hence the choice of $\boldsymbol{q}_{n-2}$ in the search forces the choice of $\boldsymbol{q}_{n-1}$ and $\boldsymbol{q}_{n}$. In all these three previous conditions, the MINVO control points are used. 

As in the optimization problem we are imposing the trajectory to
be inside the sphere $\mathcal{S}$, we also discard $\boldsymbol{q}_{l}$ if condition \ref{enu:sphere}
is not satisfied. Additionally, to keep the search computationally
tractable, we discard $\boldsymbol{q}_{l}$ if it is very close to
another $\boldsymbol{q}_{k}$ already added to $Q$ (condition \ref{enu:voxel}):
we create a voxel grid of voxel size $2\epsilon'$, and add a new control point to $Q$ only if no other point has been added
before within the same voxel. Finally, we also discard $\boldsymbol{q}_{l}$ if there
are not any feasible samples for $\boldsymbol{v}_{l}$ (condition
\ref{enu:sizeOfV}).

Then, we check if we have found all the control points and if
$\boldsymbol{q}_{n-2}$ is sufficiently close to the goal $\boldsymbol{g}$ (distance
less than $\epsilon''$). If this is the case, the control points $\boldsymbol{q}_{n-1}$
and $\boldsymbol{q}_{n}$ (which are the same as $\boldsymbol{q}_{n-2}$
due to the final stop condition) are added to the list of the corresponding
control points, and are returned together with all the separating
planes $\boldsymbol{\pi}_{ij}\;\forall i \in I, \forall j \in J$ (lines \ref{line:checkInGoal1}-\ref{line:checkInGoal2}).
If the goal has not been reached yet, we use the velocity samples
$\mathcal{M}$ to generate $\boldsymbol{q}_{l+1}$ and add them to
$Q$ (lines \ref{line:Add_samples_toQ1}-\ref{line:Add_samples_toQ-2}).
If the algorithm is not able to find a trajectory that reaches the
goal, the one found that is closest to the goal is returned (line
\ref{line:return}). 
\begin{figure}
	\begin{centering}
		\modifiedFig{\includegraphics[width=1\columnwidth]{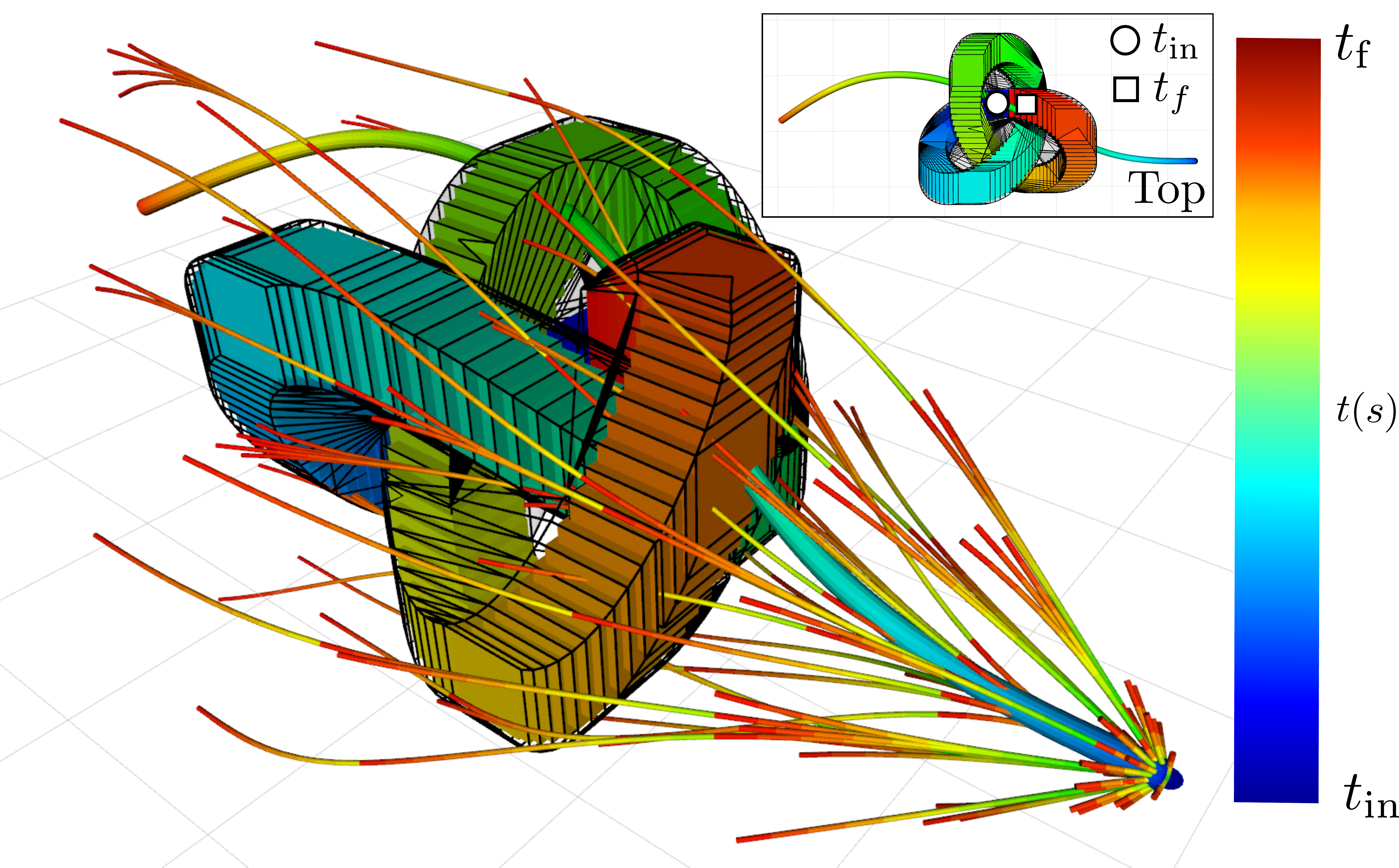}}
		\par\end{centering}
	\caption{Example of the trajectories found by the Octopus Search in an environment with a dynamic obstacle following a trefoil knot trajectory. The best trajectory found is the thickest one in the figure. \label{fig:octopus_search}}
\end{figure}
\begin{figure*}
\begin{centering}
\includegraphics[width=1\textwidth]{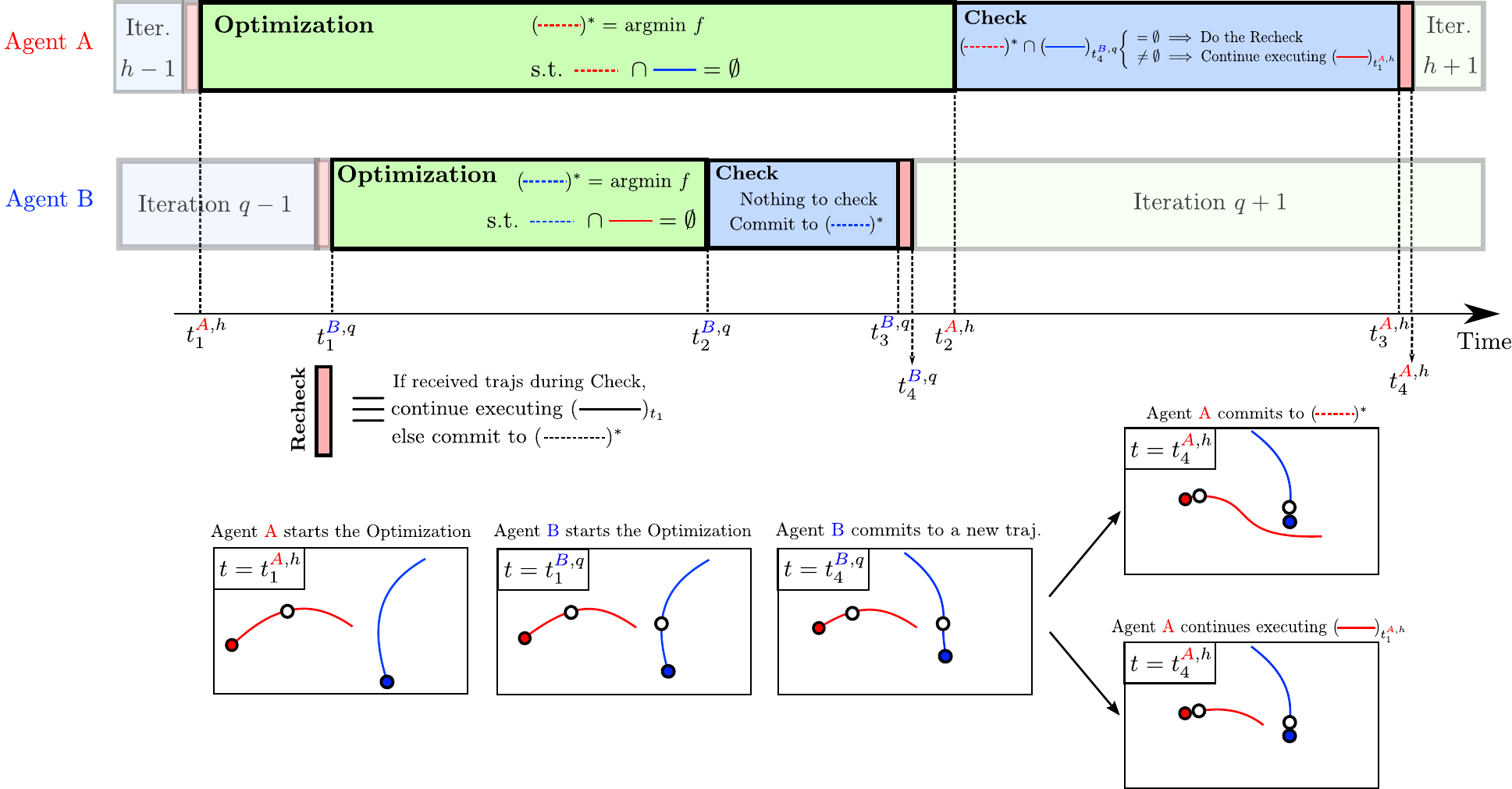}
\par\end{centering}
\caption{Deconfliction between agents. Each agent includes the trajectories other agents have committed to as constraints in the optimization. After the optimization, a collision check-recheck scheme is performed to ensure feasibility with respect to trajectories other agents have committed to while the optimization was happening. In this example, agent \textcolor{blue}{B} starts the optimization after agent \textcolor{red}{A}, but commits to a trajectory before agent  \textcolor{red}{A}. Hence, when agent \textcolor{red}{A} finishes the optimization it needs to check whether the trajectory found collides or not with the trajectory agent \textcolor{blue}{B} committed to at $t=t_{4}^{{\color{blue}B},q}$. If it collides, agent \textcolor{red}{A} will simply keep executing the trajectory available at $t=t_{1}^{{\color{red}A},h}$. If it does not collide, agent \textcolor{red}{A} will do the Recheck step to ensure no agent has committed to any trajectory during the Check period, and if this Recheck step is satisfied, agent \textcolor{red}{A} will commit to the trajectory found. \label{fig:Deconfliction}}

\vskip 1ex
\end{figure*}

Fig. \ref{fig:octopus_search} shows an example of the trajectories found by the Octopus Search algorithm in an environment with a dynamic obstacle following a trefoil knot trajectory.

\add{
\subsection{Degree of the splines}\label{subsec:degree_of_the_splines}
In this paper, we focused on the case $p=3$ (i.e., cubic splines). However, MADER could also be used with higher (or lower) order splines. For instance, one could use splines of fourth-degree polynomials (i.e., $p=4$), minimize snap (instead of jerk), and then use the corresponding MINVO polyhedron that encloses each fourth-degree interval for the obstacle avoidance constraints \cite{tordesillas2020minvo}. The reason behind the choice of cubic splines, instead of higher/lower order splines, is that cubic splines are a good trade-off between dynamic feasibility of a UAV \cite{mellinger2011minimum} and computational tractability. }

\section{Deconfliction}\label{sec:deconfliction}

\begin{figure}
\begin{centering}
\includegraphics[width=1\columnwidth]{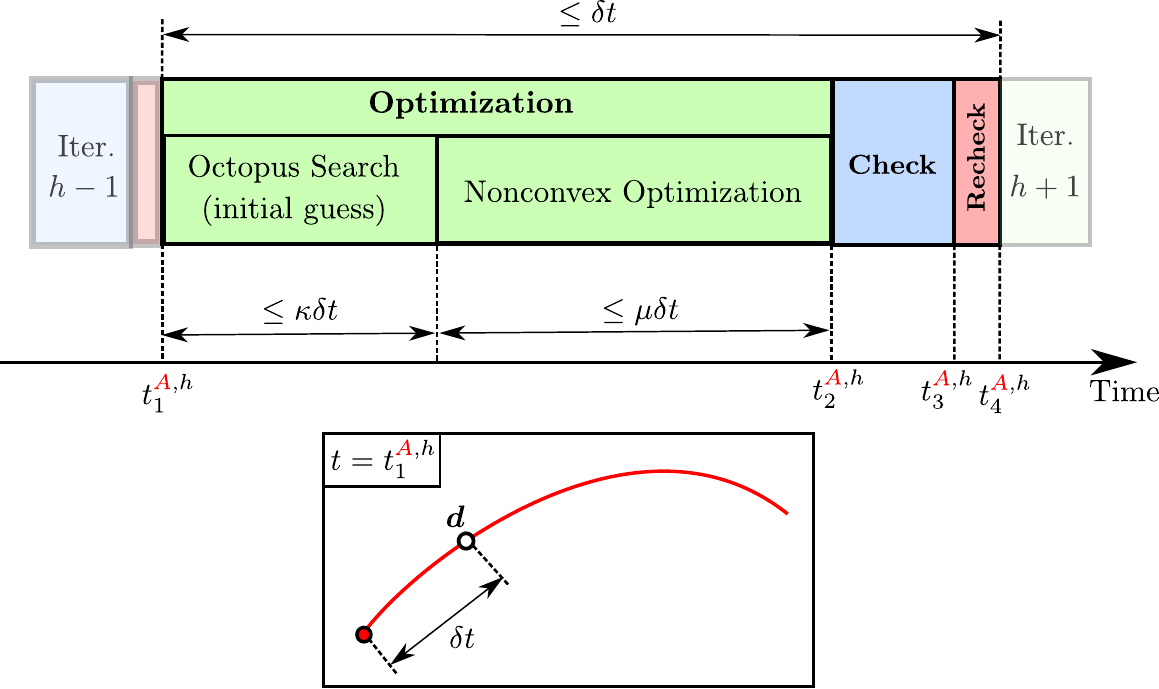}
\par\end{centering}
\caption{At $t=t_{1}^{{\color{red}A},h}$, agent \textcolor{red}{A} chooses
the point $\boldsymbol{d}$ (\tikzcircle[black,fill=white]{15pt})
along the current trajectory that it is executing, with an offset
$\delta t$ from the current position \tikzcircle[black,fill=red]{15pt}.
Then, it allocates $\kappa\delta t$ seconds to obtain an initial
guess. The closest trajectory found to $\boldsymbol{g}$ is used as
the initial guess if the search has not finished by that time. Then,
the nonconvex optimization runs for $\mu\delta t$ seconds, choosing
the best feasible solution found if no local optimum has been found
by then. $\kappa$ and $\mu$ satisfy $\kappa>0$, $\mu>0$, $\kappa+\mu<1$.\label{fig:time_selection}}

\end{figure}

To guarantee that the agents plan trajectories asynchronously while
not colliding with other agents that are also constantly replanning,
we use a deconfliction scheme divided in these three periods (see
Fig. \ref{fig:Deconfliction}): 

\definecolor{opt_color}{RGB}{203,255,182}
\definecolor{check_color}{RGB}{195,218,255}
\definecolor{recheck_color}{RGB}{255,176,176}
\begin{itemize}
\item The \fcolorbox{black}{opt_color}{\textbf{Optimization}} period happens
during $t\in(t_{1},t_{2}]$. The optimization problem will include the polyhedral outer representations of the trajectories $p_i(t),\;i\in I$ in the constraints. All the trajectories other agents commit to during the Optimization period are stored.  
\item The \fcolorbox{black}{check_color}{\textbf{Check}} happens during
$t\in(t_{2},t_{3}]$. The goal of this period is to check whether
the trajectory found in the optimization collides with the trajectories other agents have committed to during the optimization. This collision check is done by
performing feasibility tests solving the Linear Program \ref{eqn:LP_Problem}  $\forall j$
for every agent $i$ that has committed to a trajectory while the
optimization was being performed (and whose new trajectory was not
included in the constraints at $t_{1}$). A boolean flag is set to true if any other agent commits to a new trajectory during this Check period. 
\item The \fcolorbox{black}{recheck_color}{\textbf{Recheck}} period aims
at checking whether agent A has received any trajectory during
the Check period, by simply checking if the boolean flag is true or false. As this is a single Boolean comparison
in the code, it allows us to assume that no trajectories
have been published by other agents while this recheck is done, avoiding therefore an infinite loop of rechecks.
\end{itemize}

\vspace{0.2cm}
With $h$ denoting the replanning iteration of an agent A, the time allocation for each of these three periods described above is explained in Fig. \ref{fig:time_selection}: to choose the initial condition of the
iteration $h$, Agent A first chooses a point $\boldsymbol{d}$ along
the trajectory found in the iteration $h-1$, with an offset of $\delta t$
seconds from the current position \tikzcircle[black,fill=red]{15pt}. Here, $\delta t$ should be an estimate of how long iteration $h$ will take. To obtain this estimate, and similar to our previous work \cite{tordesillas2019faster}, we use
the time iteration $h-1$ took multiplied by a factor $\alpha\ge1$:
$\delta t=\alpha\left(t_{4}^{{\color{red}A},h-1}-t_{1}^{{\color{red}A},h-1}\right)$. Agent A then should finish the replanning iteration $h$ in
less than $\delta t$ seconds. To do this, we allocate a maximum runtime
of $\kappa\delta t$ seconds to obtain an initial guess, and a maximum
runtime of $\mu\delta t$ seconds for the nonconvex optimization.
Here $\kappa>0,\mu>0$ and $\kappa+\mu<1$,  
to give time for the Check and Recheck. If the Octopus Search takes
longer than $\kappa\delta t$, the trajectory found that is closest to the
goal is used as the initial guess. Similarly, if the nonconvex optimization
takes longer than $\mu\delta t$, the best feasible solution found is selected.

Fig. \ref{fig:Deconfliction} shows an example scenario with only
two agents A and B. Agent A starts its $h$-th replanning
step at $t_{1}^{{\color{red}A},h}$, and finishes the optimization at $t_{2}^{{\color{red}A},h}$.
Agent B starts its $q$-th replanning step at
$t_{1}^{{\color{blue}B},q}$. In the example
shown, agent B starts the optimization later than agent A ($t_{1}^{{\color{blue}B},q}>t_{1}^{{\color{red}A},h}$), but solves the optimization earlier than agent A ($t_{2}^{{\color{blue}B},q}<t_{2}^{{\color{red}A},h}$).
As no other agent has obtained a trajectory while agent B was optimizing,
agent B does not have to check anything, and commits directly
to the trajectory found. However, when agent A finishes the optimization
at $t_{2}^{{\color{red}A},h}>t_{4}^{{\color{blue}B},q}$, it needs
to check if the trajectory found collides with the one agent B
has committed to. If they do not collide, agent A will perform
the Recheck by ensuring that no trajectory has been published while
the Check was being performed. 

An agent will keep executing the trajectory found in the previous iteration if any of these four scenarios happens:
\begin{enumerate}
\item \label{enu:scenario1}The trajectory obtained at the end of the optimization collides with
any of the trajectories received during the Optimization. 
\item \label{enu:scenario2}The agent has received any trajectory from other agents  during the
Check period.
\item \label{enu:scenario3}No feasible solution has been found in the Optimization.
\item \label{enu:scenario4}The current iteration takes longer than $\delta t$ seconds. 
\end{enumerate}

\begin{figure}
\centering
\begin{minipage}{1.0\columnwidth}
  \centering
  \includegraphics[width=1.0\linewidth]{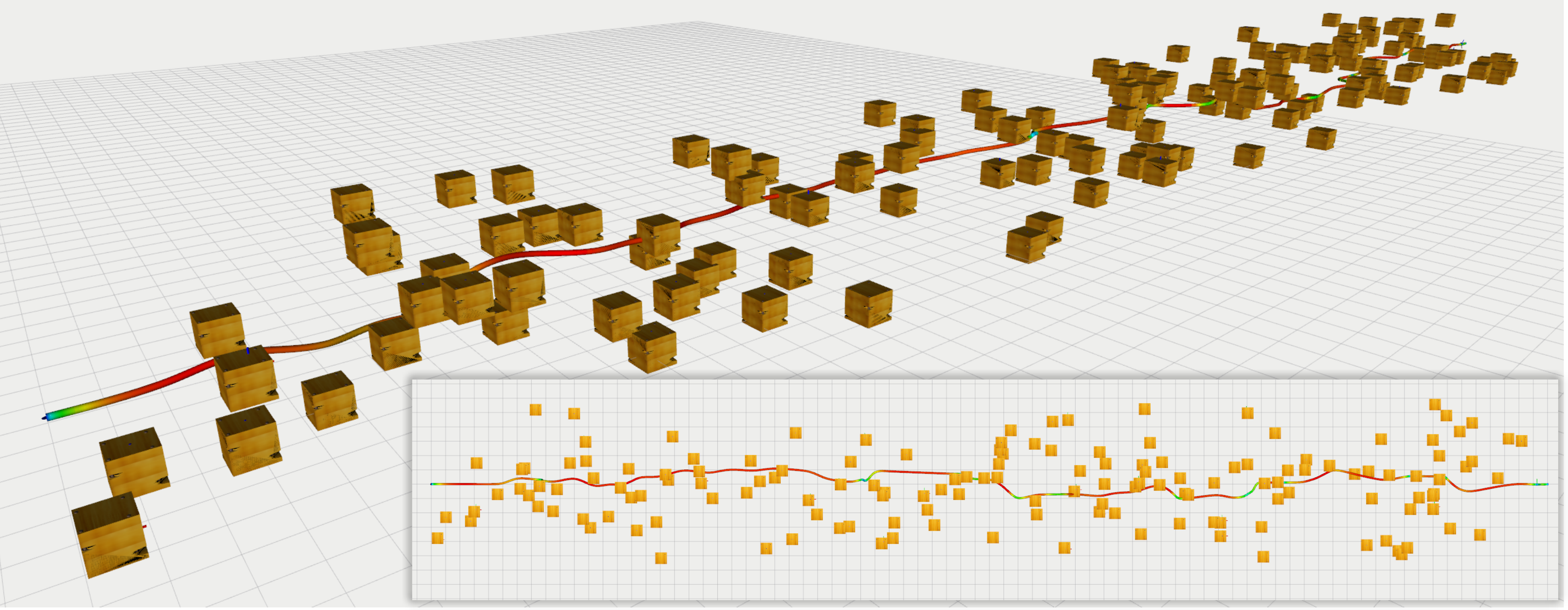}
  \captionof{figure}{Corridor environment of size $73\;\text{m}\times4\;\text{m}\times3\;\text{m}$ used for the single-agent simulation. It contains 100 randomly-deployed dynamic obstacles that follow a trefoil knot trajectory. The corridor is along the $x$ direction.}
  \label{fig:corridor}
\end{minipage}%
\hfill
\vskip 5ex
\begin{minipage}{1.0\columnwidth}
  \centering
  \includegraphics[width=1.0\linewidth]{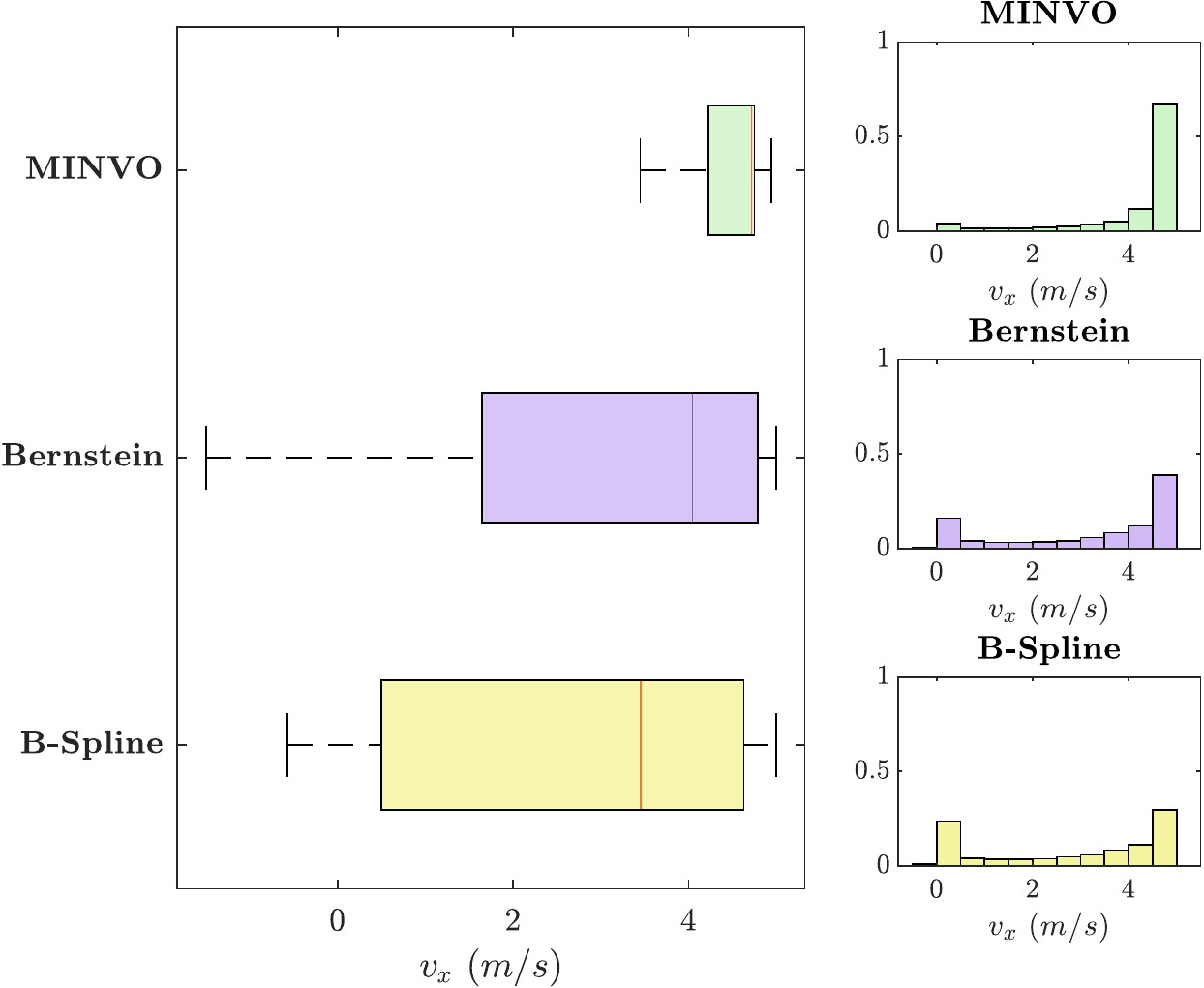}
  \captionof{figure}{Boxplots and normalized histograms of the velocity profile of $v_{x}$ in the corridor environment shown in Fig. \ref{fig:corridor}.
The velocity constraint used was $\boldsymbol{v}_{max}=5\cdot\boldsymbol{1}\;m/s$. The histograms are for all the velocities obtained across 10 different simulations.}
  \label{fig:comparisons_vel_x_sim}
\end{minipage}
\end{figure}

Under the third assumption explained in Sec. \ref{sec:Assumptions} (i.e., two agents do not commit to their trajectory at the very same time), this deconfliction scheme explained guarantees safety with respect to the other agents, which is proven as follows:
\begin{itemize}
    \item If the planning agent commits to a new trajectory in the current replanning iteration, this new trajectory is guaranteed to be collision-free because it included all the trajectories of other agents as constraints, and it was checked for collisions with respect to the trajectories other agents have committed to during the planning agent's optimization time. 
    \item If the planning agent does not commit to a new trajectory in that iteration (because one of the scenarios~\ref{enu:scenario1}--\ref{enu:scenario4} occur), it will keep executing the trajectory found in the previous iteration. This trajectory is still guaranteed to be collision-free because it was collision-free when it was obtained and other agents have included it as a constraint in any new plans that have been made recently. If the agent reaches the end of this trajectory (which has a final stop condition), the agent will wait there until it obtains a new feasible solution. In the meanwhile, all the other agents are including its position as a constraint for theirs trajectories, guaranteeing therefore safety between the agents.
\end{itemize}

\section{Results}\label{sec:Results}

\add{We now test MADER in several single-agent and multi-agent simulation environments. The computers used for the simulations are
an \texttt{AlienWare Aurora r8} desktop (for the C++ simulations of \ref{sec:ma_sim_without_obs}), a \texttt{ROG Strix GL502VM} laptop (for the Matlab simulations of Sec. \ref{sec:ma_sim_without_obs})
and a 
\texttt{general-purpose-N1} Google Cloud instance (for the simulations of Sec. \ref{subsec:singleAgentSim} and \ref{sec:MAstaticdynamic}).}

\subsection{Single-Agent simulations}\label{subsec:singleAgentSim}

To highlight the benefits of the MINVO basis with respect to the Bernstein
or B-Spline bases, we first run the algorithm proposed in a corridor-like
enviroment ($73\;\text{m}\times4\;\text{m}\times3\;\text{m}$) depicted
in Fig. \ref{fig:corridor}. that contains 100 randomly-deployed dynamic obstacles
of sizes $0.8\;\text{m}\times0.8\;\text{m}\times0.8\;\text{m}$. All the obstacles follow a trajectory
whose parametric equations are those of a trefoil knot \cite{trefoil2020}. \add{The radius of the sphere $\mathcal{S}$ used is $r=4.0$ m, and} the velocity and acceleration constraints for the UAV are $\boldsymbol{v}_{max}=5\cdot\boldsymbol{1}\;m/s$
and $\arraycolsep=2.3pt\boldsymbol{a}_{max}=\left[\begin{array}{ccc}
20 & 20 & 9.6\end{array}\right]^{T}\;m/s^{2}$. The velocity profile for $v_{x}$ is shown in Fig. \ref{fig:comparisons_vel_x_sim}. For the same given velocity constraint ($v_\text{max}=5\;m/s$), the mean velocity $v_x$ achieved by the MINVO basis is $4.15\;m/s$, higher than the ones achieved by the Bernstein and B-Spline basis ($3.23\;m/s$ and $2.79\;m/s$ respectively).

\begin{table*}
	\caption{Comparison of the number of stops and time to reach the goal in a
		corridor-like environment using different bases and with different
		number of obstacles. \label{tab:Corridor-comparison}}
	\resizebox{\textwidth}{!}{
		\begin{centering}
			\begin{tabular}{|c|c|c|c|c|c|c|c|c|c|c|}
				\cline{2-11} \cline{3-11} \cline{4-11} \cline{5-11} \cline{6-11} \cline{7-11} \cline{8-11} \cline{9-11} \cline{10-11} \cline{11-11} 
				\multicolumn{1}{c|}{} & \multicolumn{2}{c|}{\textbf{50 obstacles}} & \multicolumn{2}{c|}{\textbf{100 obstacles}} & \multicolumn{2}{c|}{\textbf{150 obstacles}} & \multicolumn{2}{c|}{\textbf{200 obstacles}} & \multicolumn{2}{c|}{\textbf{250 obstacles}}\tabularnewline
				\hline 
				\textbf{Basis} & \textbf{Stops} & \textbf{Time (s)} & \textbf{Stops} & \textbf{Time (s)} & \textbf{Stops} & \textbf{Time (s)} & \textbf{Stops} & \textbf{Time (s)} & \textbf{Stops} & \textbf{Time (s)}\tabularnewline
				\hline 
				\hline 
				\textbf{B-Spline} & $4.6\pm2.50$ & $19.61\pm1.67$ & $9.0\pm2.40$ & $26.17\pm3.04$ & $9.1\pm2.31$ & $26.97\pm3.24$ & $12.1\pm2.77$ & $32.69\pm4.61$ & $14.10\pm4.09$ & $39.81\pm9.76$\tabularnewline
				\hline 
				\textbf{Bernstein} & $5.0\pm2.0$ & $19.07\pm1.08$ & $7.4\pm2.36$ & $22.59\pm1.33$ & $6.5\pm1.72$ & $23.23\pm\boldsymbol{1.43}$ & $7.2\pm2.49$ & $24.65\pm2.54$ & $8.4\pm\boldsymbol{2.80}$ & $27.94\pm4.16$\tabularnewline
				\hline 
				\textbf{MINVO (ours)} & \textbf{$\boldsymbol{1.1\pm0.74}$} & $\boldsymbol{16.62\pm0.61}$ & \textbf{$\boldsymbol{1\pm0.94}$} & $\boldsymbol{17.55\pm0.98}$ & \textbf{$\boldsymbol{2.7\pm1.70}$} & $\boldsymbol{20.48}\pm2.26$ & \textbf{$\boldsymbol{4.2\pm1.87}$} & $\boldsymbol{22.47\pm2.19}$ & \textbf{$\boldsymbol{8.1}\pm3.45$} & $\boldsymbol{27.78\pm3.83}$\tabularnewline
				\hline 
			\end{tabular}
			\par\end{centering}
	}
\end{table*}

\begin{figure}
	\begin{centering}
		\includegraphics[width=1\columnwidth]{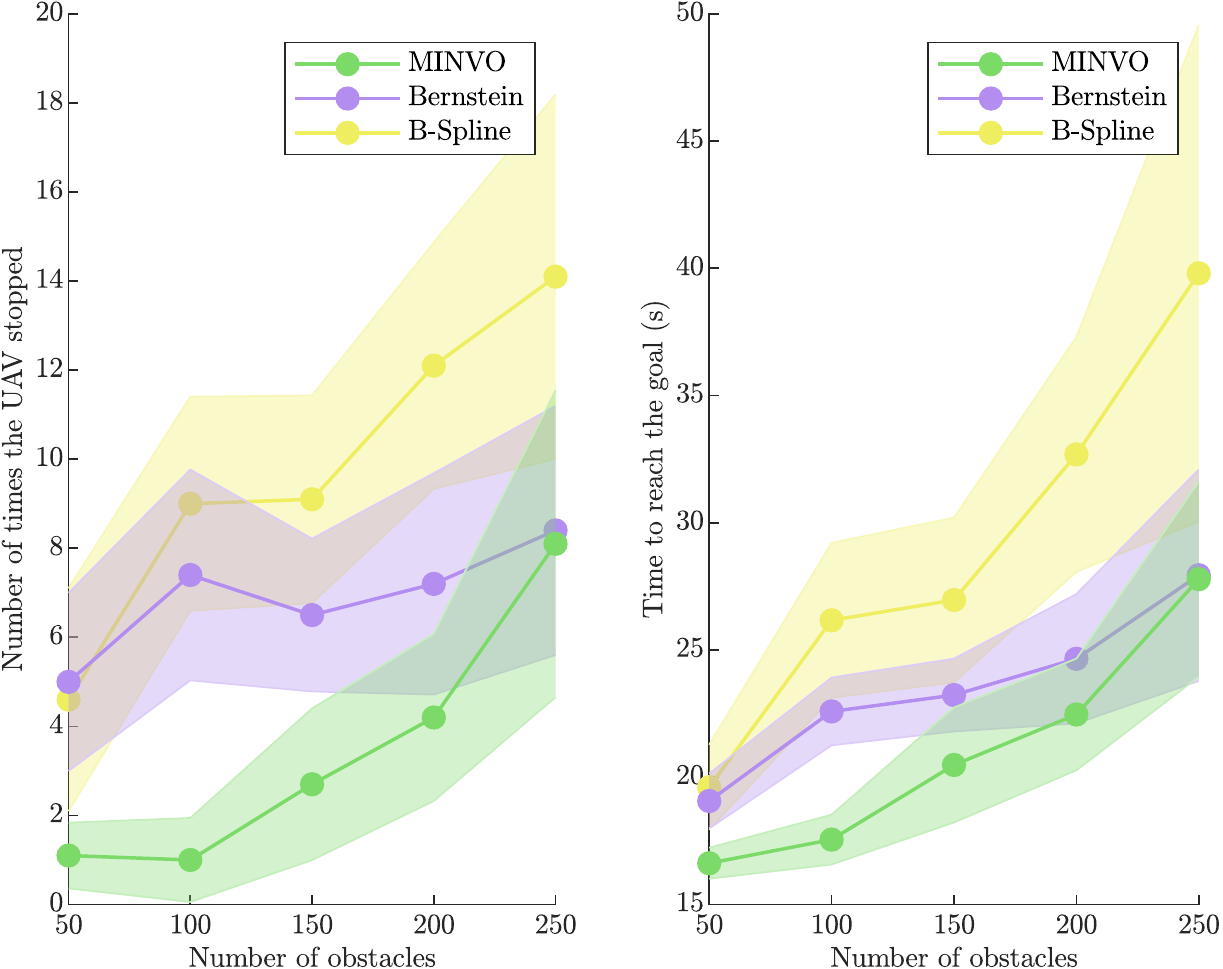}
		\par\end{centering}
	\caption{Time to reach the goal and number of times the UAV had to stop for
		different number of obstacles. 5 simulations were performed for each combination of basis (MINVO, Bernstein and B-Spline) and number of obstacles. The shaded area is the 1$\sigma$ interval, where $\sigma$ is the standard deviation. \label{fig:comparisons_varying_num_obs}}
	
	\vskip-2ex
\end{figure}

We now compare the time it takes for the UAV to reach the goal using each basis in the same corridor environment but varying the total number of obstacles (from 50 obstacles to 250 obstacles). Moreover, and as a stopping condition is not a safe condition in a world with dynamic
obstacles, we also report the number of times the UAV had to stop. The results are shown in Table \ref{tab:Corridor-comparison}
and Fig. \ref{fig:comparisons_varying_num_obs}. In terms of number
of stops, the use of the MINVO basis
achieves reductions of 86.4\% and 88.8\% with respect to the Bernstein and B-Spline bases
respectively. In terms of the time to reach the goal, the MINVO basis
achieves reductions of 22.3\% and 33.9\% compared to the Bernstein
and B-Spline bases respectively. The reason behind all these improvements
is the tighter outer polyhedral approximation of each interval of the trajectory
achieved by the MINVO basis in the velocity and position spaces.

\subsection{Multi-Agent simulations without obstacles}\label{sec:ma_sim_without_obs}

We now compare MADER with the following different state-of-the-art
algorithms:
\add{
\begin{itemize}
\item Sequential convex programming (SCP\footnote{\href{https://github.com/qwerty35/swarm_simulator}{https://github.com/qwerty35/swarm\_simulator}}, \cite{augugliaro2012generation}).
\item Relative Bernstein Polynomial approach (RBP\footnotemark[\value{footnote}], \cite{park2019efficient}).
\item Distributed model predictive control (DMPC\footnote{\label{fn:multiagent_planning}\href{https://github.com/carlosluisg/multiagent_planning}{https://github.com/carlosluisg/multiagent\_planning}}, \cite{luis2019trajectory}).
\item Decoupled incremental sequential convex programming (dec\_iSCP\footnotemark[\value{footnote}],\cite{chen2015decoupled}).
\item Search-based motion planing (\cite{liu2018towards}\footnote{\href{https://github.com/sikang/mpl_ros}{https://github.com/sikang/mpl\_ros}}),
both in its sequential version (decS\_Search) and in its non-sequential
version (decNS\_Search).
\end{itemize}
}
To classify these different algorithms, we use the following definitions:
\begin{itemize}
\item \textbf{Decentralized}: Each agent solves its own optimization problem. 
\item \textbf{Replanning:} The agents have the ability to plan several times as they fly (instead
of planning only once before starting to fly). The
algorithms with replanning are also classified according to whether
they satisfy the real-time constraint in the replanning: algorithms
that satisfy this constraint are able to replan in less than $\delta t$
or at least have a trajectory they can keep executing in case no solution
has found by then (see Fig. \ref{fig:time_selection}). Algorithms
that do not satisfy this constraint allow replanning steps longer
than $\delta t$ (which is \textbf{not} feasible in the real world),
and simulations are performed by simply having a simulation time that
runs completely independent of the real time. 
\item \textbf{Asynchronous}: The planning is triggered independently by
each agent without considering the planning status of other agents.
Examples of synchronous algorithms include the ones that trigger the optimization of all the agents at the same time or that impose that
one agent cannot plan until another agent has finished.
\item \textbf{Discretization} \textbf{for inter-agent constraints}: The
collision-free constraints between the agents are imposed only on a finite set of points of the trajectories. The discretization step will be denoted as $h$ seconds. 
\end{itemize}
In the test scenario, 8 agents are in a $8\times8$ m square, and they have to swap their positions. The velocity and
acceleration constraints used are $\boldsymbol{v}_{max}=1.7\cdot\boldsymbol{1}\;m/s$
and $a_{max}=6.2\cdot\boldsymbol{1}\;m/s^{2}$, with a drone radius of 15 cm. Moreover, \add{we define the safety ratio as $\min_{i,i^{'}}d_{min}^{i,i'}/(\rho_{i}+\rho_{i'})$~\cite{park2019efficient}, where $d_{min}^{i,i'}$ is the minimum distance over all the pairs of agents $i$ and $i'$, and $\rho_{i}$, $\rho_{i'}$
denote their respective radii. Safety is ensured if safety ratio~$>1$}. For the RBP and DMPC algorithms, the downwash
coefficient $c$ was set to $c=1$ (so that the drone is modeled as
a sphere as in all the other algorithms). 

\definecolor{RedLight}{RGB}{255,204,204}

\definecolor{GreenLight}{RGB}{204,255,204}

\newcommand{\Yes}{\cellcolor{GreenLight}\multirow{-4}{*}{Yes}}

\newcommand{\YesOne}{\cellcolor{GreenLight}Yes}

\newcommand{\NoOne}{\cellcolor{RedLight}No}

\newcommand{\No}{\cellcolor{RedLight}\multirow{-4}{*}{No}}

\newcommand{\NotApplicable}{\cellcolor{gray}\multirow{-4}{*}{ }}

\selectlanguage{english}%
\newcommand\tikzmark[2][]{   \tikz[remember picture,inner sep=\tabcolsep,outer sep=0pt,baseline=(#1.base),align=left]{\node[minimum width=0\hsize](#1){$#2$};} }

\selectlanguage{american}%
\begin{table*}
	\caption{Comparison between MADER (ours), SCP (\cite{augugliaro2012generation}), RBP (\cite{park2019efficient}), DMPC (\cite{luis2019trajectory}), dec\_iSCP (\cite{chen2015decoupled}), decS\_Search and decNS\_Search (\cite{liu2018towards}) . For SCP and MADER, the time and distance results are the mean of 5 runs, and the safety ratio is the minimum across all the runs. The test environment consists of 8 agents in a square that swap their positions without obstacles. For the algorithms that have replanning, the values in the columns of computation and execution times are $t_{1^{st}\text{start}}$~|~$t_{\text{last start}}$~|~$t_{1^{st}\text{end}}$ (see Fig. \ref{fig:Times}). The superscript $^{*}$ means the available implementation of the algorithm is in MATLAB (rest is in C++). Algorithms that have replanning but do not satisfy the real-time constraints in the replanning are denoted as Yes/No in the \textit{Replan?} column of the table.\label{tab:Comparison-between-MADER}}
	
	\resizebox{\textwidth}{!}{
		\begin{centering}
			\begin{tabular}{|c|c|c|c|>{\centering}p{1.6cm}|c|c|c|>{\centering}p{1.6cm}|c|c|}
				\hline 
				\multirow{2}{*}{\textbf{Method}} & \multirow{2}{*}{\textbf{Decentr.?}} & \multirow{2}{*}{\textbf{Replan?}} & \multirow{2}{*}{\textbf{Async.?}} & \multirow{2}{1.6cm}{\textbf{Without discretization?}} & \multicolumn{3}{c|}{\textbf{Time (s)}} & \multirow{2}{1.6cm}{\textbf{Total Flight Distance (m)}} & \multicolumn{2}{c|}{\textbf{Safety}}\tabularnewline
				\cline{6-8} \cline{7-8} \cline{8-8} \cline{10-11} \cline{11-11} 
				&  &  &  &  & \textbf{Computation} & \textbf{Execution} & \textbf{Total} &  & \textbf{Safe?} & \textbf{Safety ratio}\tabularnewline
				\hline 
				\hline 
				SCP, $h_{\text{SCP}}=0.3$ s & \cellcolor{RedLight} & \cellcolor{RedLight} & \cellcolor{RedLight} & \cellcolor{RedLight} & 1.150 & 7.215 & 8.366 & 77.144 & \NoOne & 0.142\tabularnewline
				SCP, $h_{\text{SCP}}=0.25$ s & \cellcolor{RedLight} & \cellcolor{RedLight} & \cellcolor{RedLight} & \cellcolor{RedLight} & 3.099 & 7.855 & 10.954 & 101.733 & \NoOne & 0.149\tabularnewline
				SCP, $h_{\text{SCP}}=0.2$ s & \cellcolor{RedLight} & \cellcolor{RedLight} & \cellcolor{RedLight} & \cellcolor{RedLight} & 8.741 & 7.855 & 16.596 & 90.917 & \NoOne & 0.335\tabularnewline
				SCP, $h_{\text{SCP}}=0.17$ s & \No & \No & \No & \No & 37.375 & 9.775 & 47.150 & 77.346 & \NoOne & 0.685\tabularnewline
				\hline 
				RBP, batch\_size=1 & \cellcolor{RedLight} & \cellcolor{RedLight} & \cellcolor{RedLight} & \cellcolor{GreenLight} & 0.228 & 15.623 & 15.851 & 90.830 & \YesOne & 1.055\tabularnewline
				RBP, batch\_size=2 & \cellcolor{RedLight} & \cellcolor{RedLight} & \cellcolor{RedLight} & \cellcolor{GreenLight} & 0.236 & 15.623 & 15.859 & 91.789 & \YesOne & 1.075\tabularnewline
				RBP, batch\_size=4 & \cellcolor{RedLight} & \cellcolor{RedLight} & \cellcolor{RedLight} & \cellcolor{GreenLight} & 0.277 & 14.203 & 14.480 & 92.133 & \YesOne & 1.023\tabularnewline
				RBP, no batches & \No & \No & \No & \Yes & 0.461 & 14.203 & 14.664 & 93.721 & \YesOne & 1.057\tabularnewline
				\hline 
				DMPC$^{*}$, $h_{\text{DMPC}}=0.45$ s  & \cellcolor{GreenLight} & \cellcolor{RedLight} & \cellcolor{RedLight} & \cellcolor{RedLight} & 4.952 & 23.450 & 28.402 & 79.650 & \NoOne & 0.683\tabularnewline
				DMPC$^{*}$, $h_{\text{DMPC}}=0.36$ s  & \cellcolor{GreenLight} & \cellcolor{RedLight} & \cellcolor{RedLight} & \cellcolor{RedLight} & 4.350 & 20.710 & 25.060 & 97.220 & \NoOne & 0.914\tabularnewline
				DMPC$^{*}$, $h_{\text{DMPC}}=0.3$ s  & \cellcolor{GreenLight} & \cellcolor{RedLight} & \cellcolor{RedLight} & \cellcolor{RedLight} & 4.627 & 18.430 & 23.057 & 78.580 & \YesOne & 1.015\tabularnewline
				DMPC$^{*}$, $h_{\text{DMPC}}=0.25$ s  & \Yes & \No & \No & \No & 3.796 & 15.980 & 19.776 & 79.590 & \NoOne & 0.824\tabularnewline
				\hline 
				dec\_iSCP$^{*}$, $h_{\text{iSCP}}=0.4$ s  & \cellcolor{GreenLight} & \cellcolor{RedLight} & \cellcolor{RedLight} & \cellcolor{RedLight} & 2.631 & 13.130 & 15.761 & 102.320 & \NoOne & 0.017\tabularnewline
				dec\_iSCP$^{*}$, $h_{\text{iSCP}}=0.3$ s  & \cellcolor{GreenLight} & \cellcolor{RedLight} & \cellcolor{RedLight} & \cellcolor{RedLight} & 4.276 & 15.470 & 19.746 & 97.770 & \NoOne & 0.550\tabularnewline
				dec\_iSCP$^{*}$, $h_{\text{iSCP}}=0.2$ s  & \cellcolor{GreenLight} & \cellcolor{RedLight} & \cellcolor{RedLight} & \cellcolor{RedLight} & 9.639 & 14.060 & 23.699 & 95.790 & \NoOne & 0.917\tabularnewline
				dec\_iSCP$^{*}$, $h_{\text{iSCP}}=0.15$ s  & \Yes & \No & \No & \No & 14.138 & 14.060 & 28.198 & 102.030 & \NoOne & 0.906\tabularnewline
				\hline 
				decS\_Search, $u=2\;m/s^{3}$ & \cellcolor{GreenLight} & \cellcolor{RedLight} & \cellcolor{RedLight} & \cellcolor{GreenLight} & 15.336 & 6.491 & 21.827 & 79.354 & \YesOne & 1.337\tabularnewline
				decS\_Search, $u=3\;m/s^{3}$ & \cellcolor{GreenLight} & \cellcolor{RedLight} & \cellcolor{RedLight} & \cellcolor{GreenLight} & 7.772 & 6.993 & 14.764 & 80.419 & \YesOne & 1.768\tabularnewline
				decS\_Search, $u=4\;m/s^{3}$ & \cellcolor{GreenLight} & \cellcolor{RedLight} & \cellcolor{RedLight} & \cellcolor{GreenLight} & 34.557 & 9.491 & 44.048 & 83.187 & \YesOne & 1.491\tabularnewline
				decS\_Search, $u=5\;m/s^{3}$ & \Yes & \No & \No & \Yes & 3.104 & 8.491 & 11.595 & 80.804 & \YesOne & 1.474\tabularnewline
				\hline 
				decNS\_Search, $u=2\;m/s^{3}$ & \cellcolor{GreenLight} & \multirow{4}{*}{\hspace{-2.5ex}\tikzmark[a]{\raisebox{2.7ex}{Yes}\raisebox{-3ex}{\hspace{3ex}No}}\hspace{-2.5ex}} & \cellcolor{RedLight} & \cellcolor{GreenLight} & \multicolumn{2}{c|}{0.021 | 0.233 | 33.711} & 34.288 & 80.116 & \YesOne & 1.416\tabularnewline
				decNS\_Search, $u=3\;m/s^{3}$ & \cellcolor{GreenLight} &  & \cellcolor{RedLight} & \cellcolor{GreenLight} & \multicolumn{2}{c|}{0.003 | 0.058 | 12.810} & 13.346 & 79.752 & \YesOne & 1.150\tabularnewline
				decNS\_Search, $u=4\;m/s^{3}$ & \cellcolor{GreenLight} &  & \cellcolor{RedLight} & \cellcolor{GreenLight} & \multicolumn{2}{c|}{0.030 | 0.600 | 53.824} & 53.899 & 79.752 & \YesOne & 1.502\tabularnewline
				decNS\_Search, $u=5\;m/s^{3}$ & \Yes &  & \No & \Yes & \multicolumn{2}{c|}{0.002 | 0.051 | 9.096} & 9.169 & 88.753 & \YesOne & 1.335\tabularnewline
				\hline 
				\textbf{\rule{0pt}{10pt}MADER (ours)} & \YesOne & \YesOne & \YesOne & \YesOne & \multicolumn{2}{c|}{0.550 | 2.468 | 7.975} & 10.262 & 82.487 & \YesOne & 1.577\tabularnewline
				\hline 
			\end{tabular}
			\par\end{centering}
		\begin{tikzpicture}[remember picture,overlay]
			\path[fill=green,opacity=0.2](a.north east)--(a.south west) -- (a.north west) -- cycle;
			\path[fill=red,opacity=0.2](a.south west)--(a.south east) -- (a.north east) -- cycle;
		\end{tikzpicture}
		
	}
\end{table*}

\begin{figure}
\begin{centering}
\includegraphics[width=1\columnwidth]{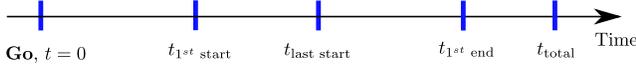}
\par\end{centering}
\caption{Time notation for the algorithms that have replanning. \label{fig:Times}}

\vskip-2ex
\end{figure}

The results obtained, together with the classification of each algorithm, are shown in Table \ref{tab:Comparison-between-MADER}.
For the algorithms that replan as they fly, we show the following
times (see Fig. \ref{fig:Times}): $t_{1^{st}\text{start}}$ (earliest
time a UAV starts flying), $t_{\text{last start}}$ (latest time a
UAV starts flying), $t_{1^{st}\text{end}}$ (earliest time a UAV reaches
the goal) and the total time $t_{\text{total}}$ (time when all the UAVs have reached
their goals). Note that the algorithm decNS\_Search is synchronous (\href{https://github.com/sikang/mpl_ros/blob/011c1ae3a47fe08c497821e45c2aabc44ddda9d7/mpl_test_node/src/robot_team.hpp\#L63}{link}), and it does not satisfy the real-time constraints in the replanning iterations (\href{https://github.com/sikang/mpl_ros/blob/master/mpl_test_node/src/multi_robot_node.cpp\#L108}{link}). Several conclusions can be drawn from Table \ref{tab:Comparison-between-MADER}:
\begin{itemize}
\item Algorithms that use discretization to impose inter-agent constraints
are in general not safe due to the fact that the constraints may
not be satisfied between two consecutive discretization points. A smaller discretization
step may solve this, but at the expense of very high computation times. 
\item Compared to the centralized solution that generates safe trajectories (RBP), MADER achieves a
shorter overall flight distance. The total time of MADER is
also shorter than the one of RBP.
\item Compared to decentralized algorithms (DMPC, dec\_iSCP, decS\_Search
and decNS\_Search), MADER is the one with the shortest total time, except for the case of decNS\_Search with $u=5\;m/s^{3}$.
However, for this case the flight distance achieved by MADER is $6.3$ m shorter. Moreover, MADER is asynchronous and satisfies the real-time
constraints in the replanning, while decNS\_Search does not.
\item From all the algorithms shown in Table \ref{tab:Comparison-between-MADER},
MADER is the only algorithm that is decentralized, has replanning, satisfies the real-time
constraints in the replanning and is asynchronous. 
\end{itemize}

\begin{figure*}[tp]
\begin{centering}
\subfloat[4 agents]{

\includegraphics[width=1.03\columnwidth]{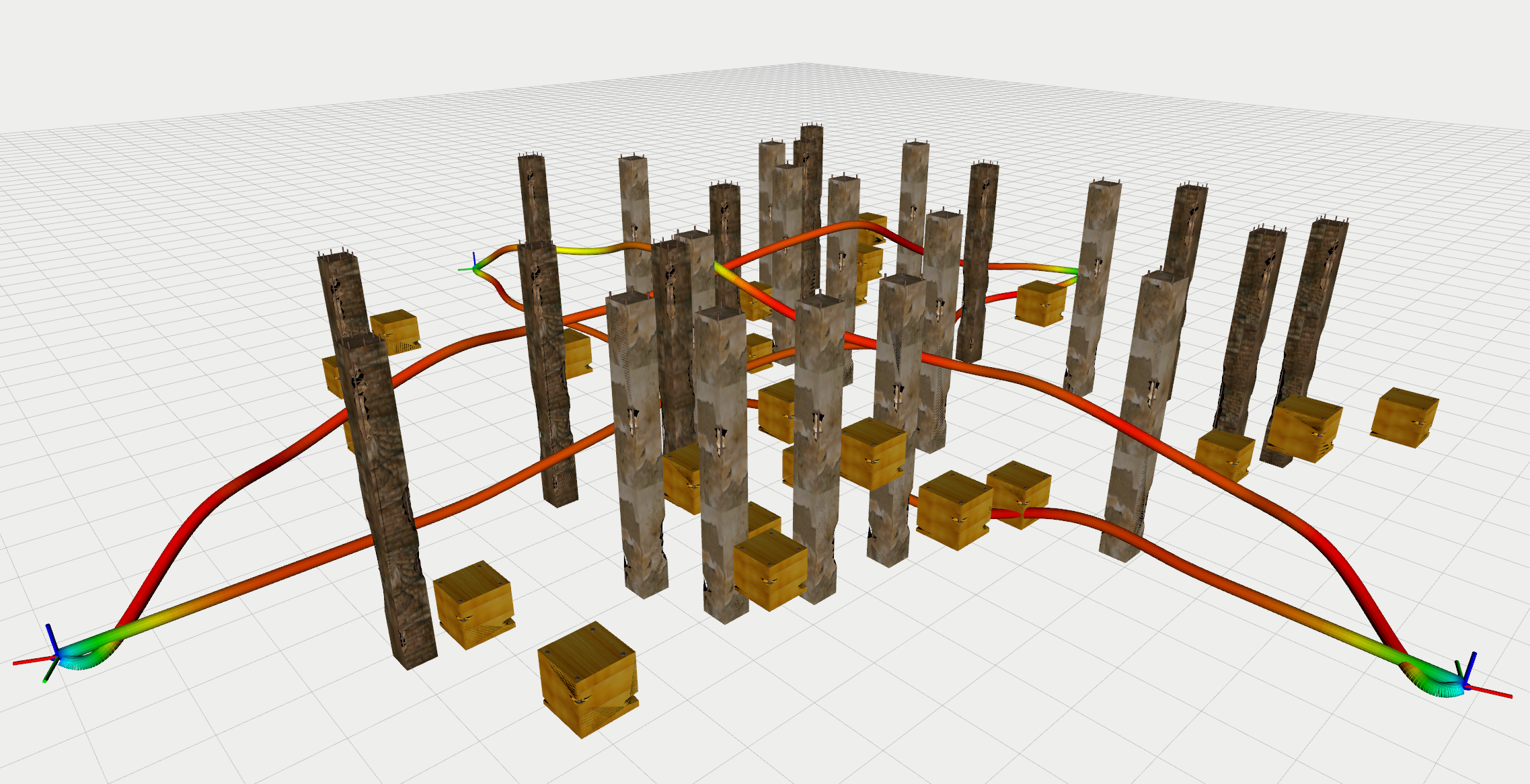}}\hfill{}\subfloat[8 agents]{\includegraphics[width=0.97\columnwidth]{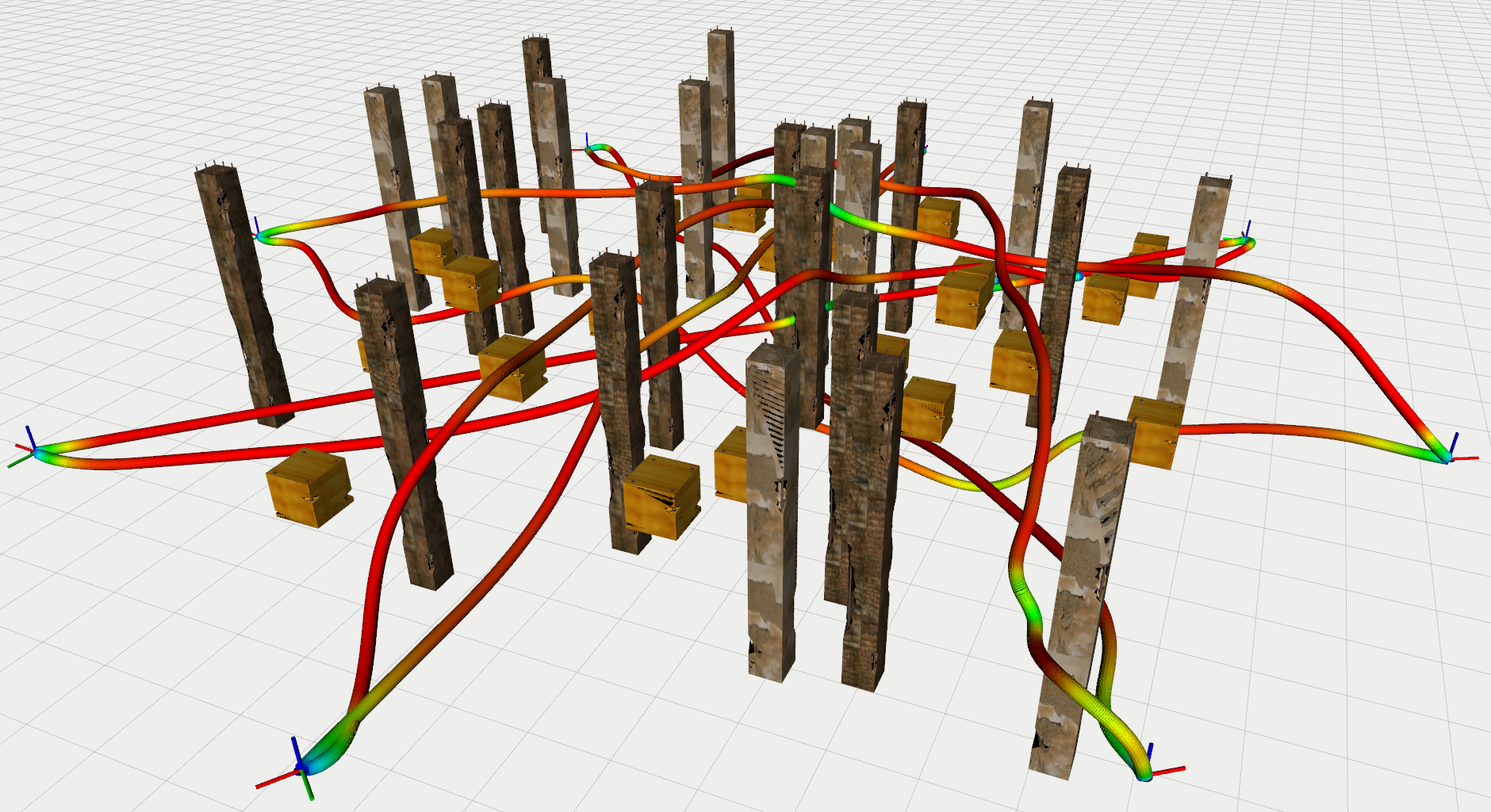}}
\par\end{centering}
\begin{centering}
\subfloat[16 agents]{\includegraphics[width=1\columnwidth]{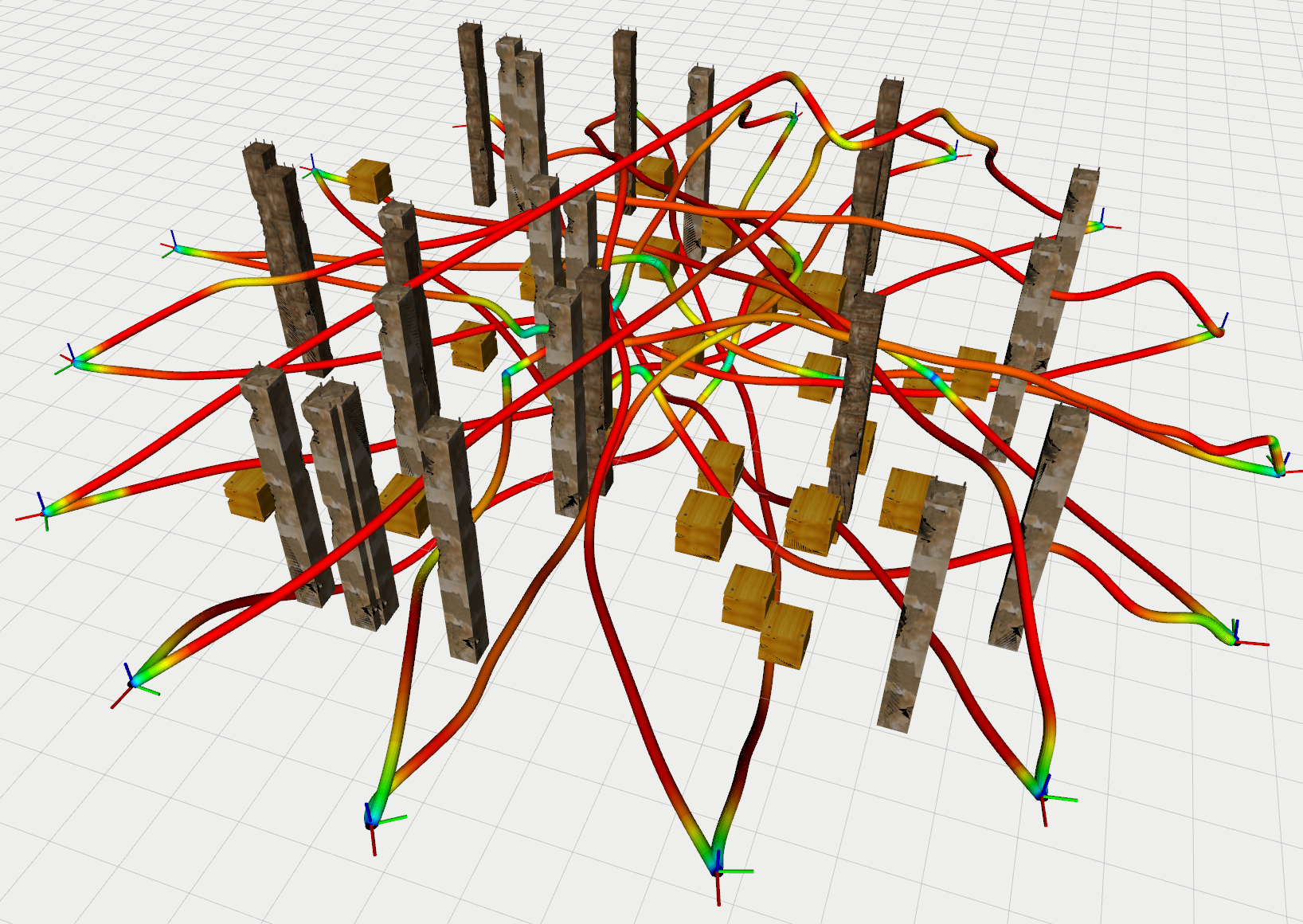}}
\par\end{centering}
\caption{Results for the circle environment, that contains 25 static obstacles (pillars) and 25 dynamic obstacles (boxes). The case with 32 agents is shown in Fig. \ref{fig:Circle32}. \label{fig:MADER-circle}}

\vskip-2ex
\end{figure*}

\begin{figure*}[htbp]
\centering
\subfloat[4 agents.]{\includegraphics[height=0.3\textwidth]{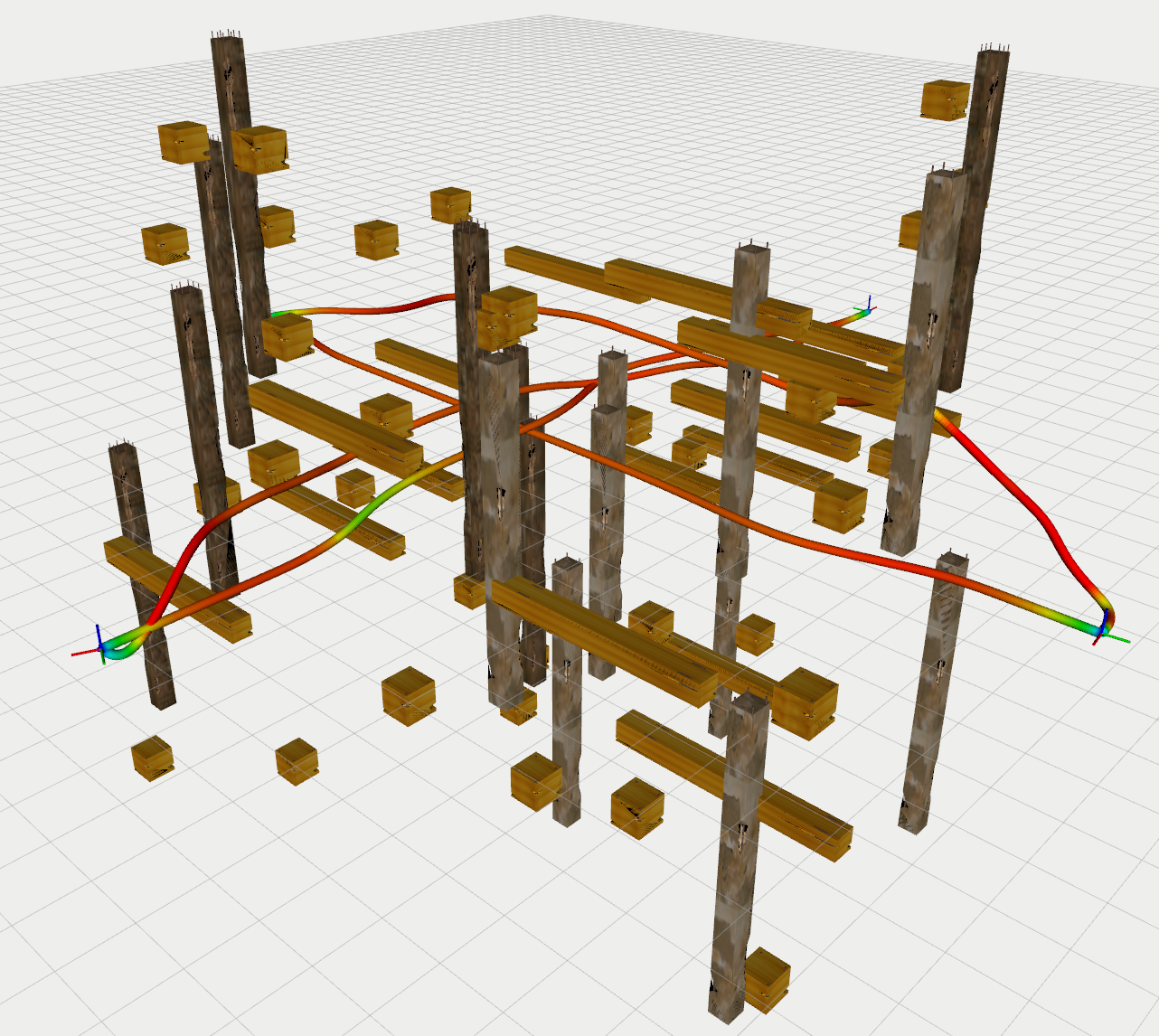}}\label{fig:1a}
\subfloat[8 agents.] {\includegraphics[height=0.3\textwidth]{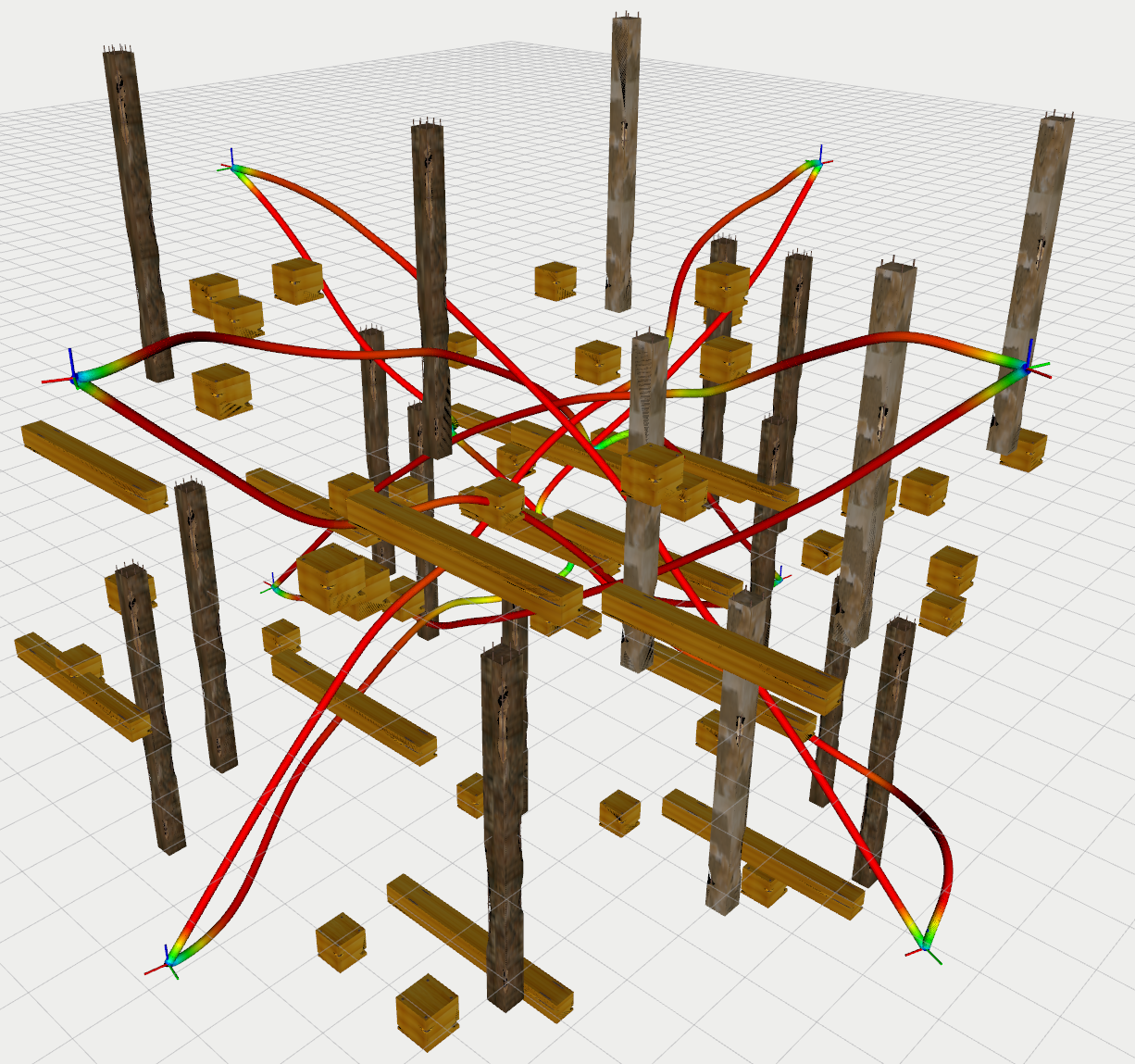}}\label{fig:1b}
\subfloat[16 agents.]{\includegraphics[height=0.3\textwidth]{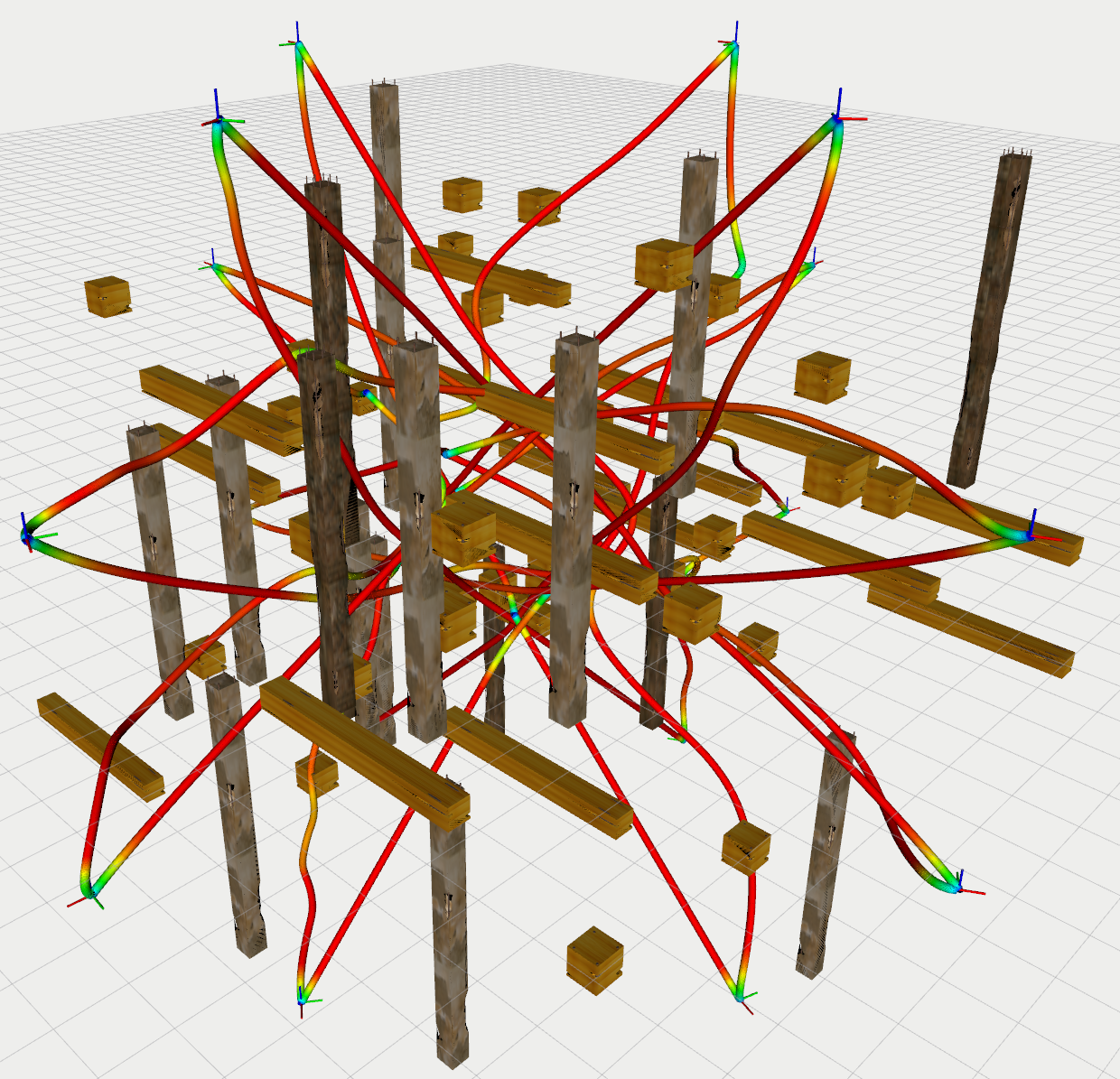}}\label{fig:1c}
\caption{Results for the sphere environment, that contains 18 static obstacles (pillars) and 52 dynamic obstacles  (boxes and horizontal poles). The case with 32 agents is shown in Fig. \ref{fig:Sphere32}. } \label{fig:MADER-sphere}
\end{figure*}

\begin{table*}[t]
\caption{Results for MADER in the circle and sphere environments.\label{tab:Results-for-MADER}}

\resizebox{\textwidth}{!}{
\begin{centering}
\begin{tabular}{|c|c|c|c|c|c|>{\centering}b{2cm}|>{\centering}b{2cm}|>{\centering}b{2cm}|}
\hline 
\multirow{2}{*}{\textbf{Environment}} & \multirow{2}{*}{\textbf{Num of Agents}} & \multicolumn{4}{c|}{\textbf{Time (s)}} & \multirow{2}{2cm}{\textbf{Flight Distance per agent (m)}} & \multirow{2}{2cm}{\textbf{Safety ratio between agents}} & \multirow{2}{2cm}{\textbf{Number of stops per agent}}\tabularnewline
\cline{3-6} \cline{4-6} \cline{5-6} \cline{6-6} 
 &  & $t_{1^{st}\text{start}}$  & $t_{\text{last start}}$ & $t_{1^{st}\text{end}}$ & \textbf{Total} &  &  & \tabularnewline
\hline 
\hline 
\multirow{4}{*}{\textbf{Circle} } & 4 & 0.339 & 0.563 & 10.473 & 11.403 & 21.052 & 5.444 & 0.000\tabularnewline
\cline{2-9} \cline{3-9} \cline{4-9} \cline{5-9} \cline{6-9} \cline{7-9} \cline{8-9} \cline{9-9} 
 & 8 & 0.404 & 0.559 & 7.544 & 11.108 & 21.025 & 1.959 & 0.000\tabularnewline
\cline{2-9} \cline{3-9} \cline{4-9} \cline{5-9} \cline{6-9} \cline{7-9} \cline{8-9} \cline{9-9} 
 & 16 & 0.300 & 0.764 & 7.773 & 13.972 & 21.737 & 4.342 & 0.188\tabularnewline
\cline{2-9} \cline{3-9} \cline{4-9} \cline{5-9} \cline{6-9} \cline{7-9} \cline{8-9} \cline{9-9} 
 & 32 & 0.532 & 1.195 & 9.092 & 18.820 & 22.105 & 1.834 & 1.500\tabularnewline
\hline 
\multirow{4}{*}{\textbf{Sphere} } & 4 & 0.452 & 0.584 & 10.291 & 11.124 & 20.827 & 4.242 & 0.000\tabularnewline
\cline{2-9} \cline{3-9} \cline{4-9} \cline{5-9} \cline{6-9} \cline{7-9} \cline{8-9} \cline{9-9} 
 & 8 & 0.425 & 0.618 & 9.204 & 12.561 & 21.684 & 1.903 & 0.125\tabularnewline
\cline{2-9} \cline{3-9} \cline{4-9} \cline{5-9} \cline{6-9} \cline{7-9} \cline{8-9} \cline{9-9} 
 & 16 & 0.363 & 0.845 & 8.909 & 13.175 & 21.284 & 1.905 & 0.125\tabularnewline
\cline{2-9} \cline{3-9} \cline{4-9} \cline{5-9} \cline{6-9} \cline{7-9} \cline{8-9} \cline{9-9} 
 & 32 & 0.357 & 1.725 & 9.170 & 18.275 & 22.284 & 1.155 & 1.000\tabularnewline
\hline 
\end{tabular}
\par\end{centering}
}
\end{table*}

\add{For MADER, and measured on the simulation environment used in this Section, each UAV performs on average 12 successful replans before reaching the goal. On average, the check step takes $\approx 2.87$ ms, the recheck step takes $\approx 0.034$ $\mu$s, and the total replanning time is $\approx 199.6$ ms. Approximately half of this replanning time is allocated to find the initial guess (i.e., $\kappa=0.5$).}

\subsection{Multi-Agent simulations with static and dynamic obstacles}\label{sec:MAstaticdynamic}

We now test MADER in multi-agent environments that have also static
and dynamic obstacles. For this set of experiments, we use $\boldsymbol{\alpha}_{j}=\boldsymbol{\beta}_{j}=3\cdot\boldsymbol{1}$
cm, $\gamma_{j}=0.1$ s $\forall j$, \add{$r=4.5$ m (radius of the sphere $\mathcal{S}$)}, and a drone radius of $5$
cm. We test MADER in the following two environments:
\begin{itemize}
\item \textbf{Circle environment}: the UAVs start in a circle formation
and have to swap their positions while flying in a world with 25 static
obstacles of size $0.4\;\text{m}\times8\;\text{m}\times0.4\;\text{m}$ and 25 dynamic obstacles
of size $0.6\;\text{m}\times0.6\;\text{m}\times0.6\;\text{m}$ following a trefoil knot trajectory
\cite{trefoil2020}. The radius of the circle the UAVs start from is $10$ m. 
\item \textbf{Sphere environment}: the UAVs start in a sphere formation
and have to swap their positions while flying in a world with 18 static
obstacles of size $0.4\;\text{m}\times8\;\text{m}\times0.4\;\text{m}$, 17 dynamic obstacles
of size $0.4\;\text{m}\times4\;\text{m}\times0.4\;\text{m}$ (moving in $z$) and 35 dynamic obstacles
of size $0.6\;\text{m}\times0.6\;\text{m}\times0.6\;\text{m}$ following a trefoil knot trajectory.
The radius of the sphere the UAVs start from is $10$ m. 
\end{itemize}
The results can be seen in Table \ref{tab:Results-for-MADER} and
in Figs. \ref{fig:32-agents-circle-sphere}, \ref{fig:MADER-circle}
and \ref{fig:MADER-sphere}. All the safety ratios between the agents
are $>1$, and the flight distances achieved (per agent) are approximately
$21.5$ m. \add{With respect to the number of stops, none of the UAVs had to stop in the circle environment with 4 and 8 agents and in the sphere environment with 4 agents. For the circle environment with 16 and 32 agents, each UAV stops (on average) 0.188 and 1.5 times respectively. For the sphere environment with 8, 16, and 32 agents, each UAV stops (on average) 0.125, 0.125, and 1.0 times respectively. 
\\	
Note also that only the two agents that have been the closest are the ones that determine the actual value of the safety ratio, while the other agents do not contribute to this value. This means that, while the safety ratio is likely to decrease with the number of agents, a monotonic decrease of the safety ratio with respect to the number of agents is not strictly required. }

\section{Conclusions}\label{sec:Conclusions}

This work presented MADER, a decentralized and asynchronous planner
that handles static obstacles, dynamic obstacles and other agents.
By using the MINVO basis, MADER obtains outer polyhedral representations of the trajectories that are 2.36 and 254.9 times smaller than the volumes
achieved using the Bernstein and B-Spline bases. To ensure non-conservative,
collision-free constraints with respect to other obstacles and agents,
MADER includes as decision variables the planes that separate each pair of outer polyhedral representations. 
Safety with respect to other agents is guaranteed in a decentralized and asynchronous way by including their committed trajectories as constraints in the optimization and then executing a collision check-recheck scheme. 
Extensive simulations in dynamic multi-agent
environments have highlighted the improvements of MADER with respect to
other state-of-the-art algorithms in terms of number of stops, computation/execution time and flight distance. \add{Future work includes adding perception-aware and risk-aware terms in the objective function, as well as hardware experiments.}

\section{Acknowledgements}

The authors would like to thank Pablo Tordesillas, Michael Everett,
Dr.\ Kasra Khosoussi, Andrea Tagliabue and Parker Lusk for helpful insights
and discussions. Thanks also to 	
Andrew Torgesen, Jeremy Cai, and 	
Stewart Jamieson for their comments on the paper.
Research funded in part by Boeing Research \& Technology.

\FloatBarrier

\selectlanguage{english}%
\bibliographystyle{IEEEtran}
\addcontentsline{toc}{section}{\refname}\bibliography{bibliography}

\selectlanguage{american}%

\begin{IEEEbiography}[{\includegraphics[width=1in,height=1.25in,clip,keepaspectratio]{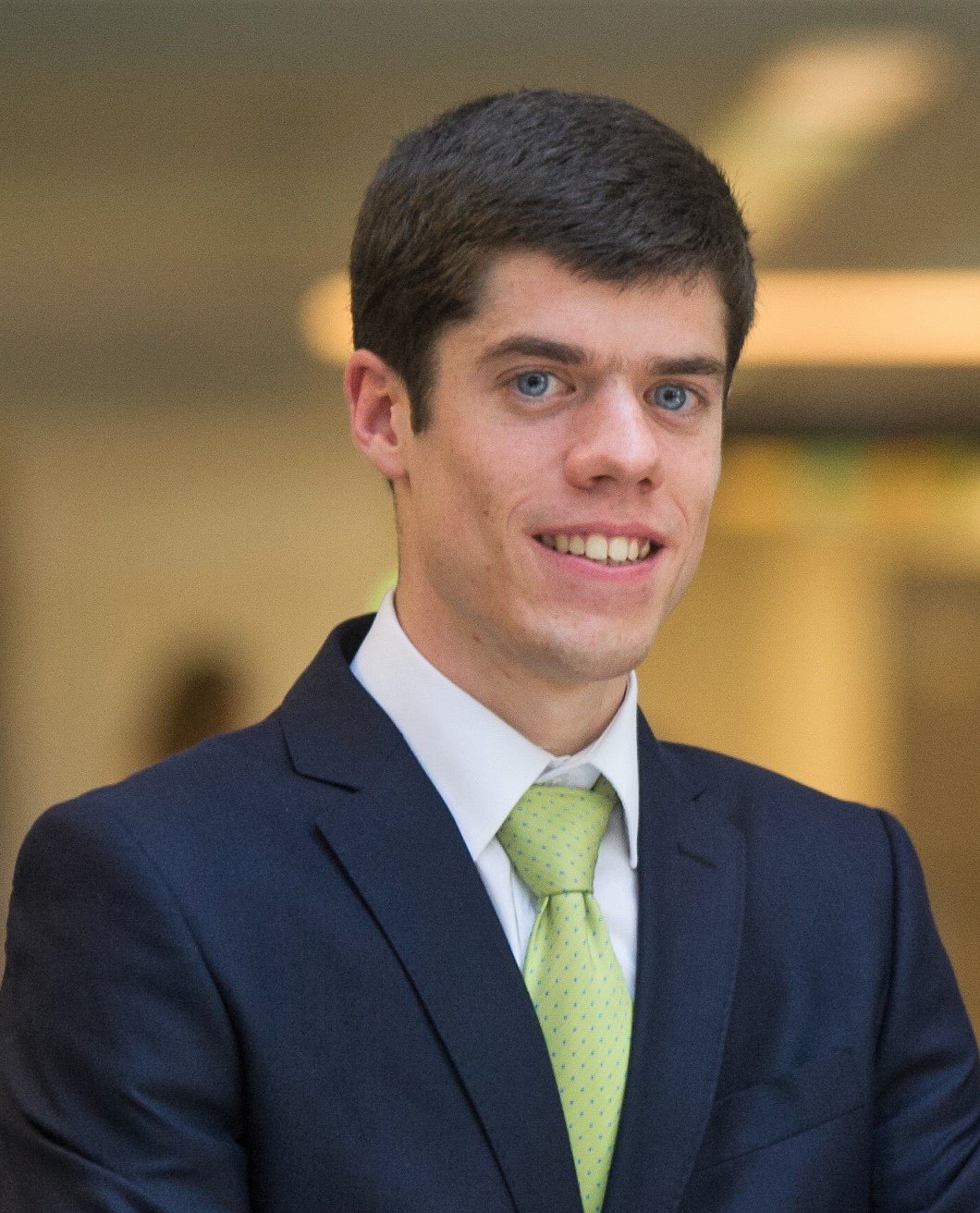}}]{Jesus Tordesillas}
(Student Member, IEEE)
received the B.S. and M.S. degrees in Electronic engineering and Robotics from the Technical University of
Madrid (Spain) in 2016 and 2018 respectively. He then received his M.S. in Aeronautics and Astronautics from MIT in 2019. He is currently pursuing the PhD degree with the Aeronautics and
Astronautics Department, as a member of the Aerospace Controls Laboratory (MIT) under the supervision of Jonathan P. How. 
His research interests include path planning for UAVs in unknown environments and optimization. His work was a finalist for the Best Paper Award on Search and Rescue Robotics in IROS 2019.
\end{IEEEbiography}

\begin{IEEEbiography}[{\includegraphics[width=2.9in,height=1.25in,clip,keepaspectratio]{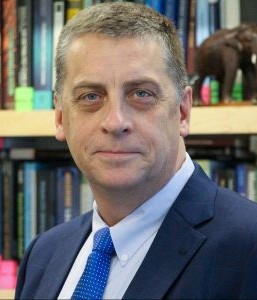}}]{Jonathan P. How}
(Fellow, IEEE) received the
B.A.Sc. degree from the University of Toronto (1987), and the S.M. and Ph.D. degrees in aeronautics and astronautics from MIT (1990 and 1993). Prior to joining MIT in 2000,
he was an Assistant Professor at Stanford University. He is currently the
Richard C. Maclaurin Professor of aeronautics and astronautics at MIT. \add{Some of his awards include the IEEE CSS Distinguished Member Award (2020), AIAA Intelligent Systems Award (2020), 
IROS Best Paper Award on Cognitive Robotics (2019), and the AIAA Best
Paper in Conference Awards (2011, 2012, 2013). 
He was the Editor-in-chief of IEEE Control Systems Magazine (2015--2019), is a Fellow of AIAA, and 
was elected to the National Academy of Engineering in 2021.}
\end{IEEEbiography}

\begin{center}
\par\end{center}

\end{document}